%% file: main.tex
\PassOptionsToPackage{table}{xcolor}
\documentclass[]{ummreport}
\usepackage[utf8]{inputenc}
\usepackage{url}
\usepackage{amsfonts,amsmath,amssymb}
\usepackage{nicefrac}
\usepackage{colortbl}
\usepackage{adjustbox}
\usepackage{tikz}
\usepackage{algorithm}
\usepackage{algorithmic}
\usepackage{pifont}
\usepackage{array}
\usepackage{tabularx}
\usepackage{xspace}
\usepackage{enumitem}
\setlist{nosep,leftmargin=*}
\makeatletter
\def\ttl@gobblecontents#1#2#3#4{\ignorespaces}
\makeatother
\usepackage{titletoc}
\titlecontents{lsection}[1.8em]{\addvspace{3pt}\bfseries\sffamily}{\contentslabel{1.8em}}{\hspace*{-1.8em}}{\titlerule*[0.6pc]{.}\contentspage}
\titlecontents{lsubsection}[4.2em]{}{\contentslabel{2.4em}}{\hspace*{-2.4em}}{\titlerule*[0.6pc]{.}\contentspage}
\definecolor{oursbg}{HTML}{F6E2A0}
\definecolor{apibg}{HTML}{DCEBFB}
\definecolor{trainbg}{HTML}{FADBDB}

\newcommand{\method}{\mbox{CaPtain}}
\newcommand{\critic}{physics critic}
\newcommand{\model}{\mbox{CaPtain-4B}}
\newcommand{\harness}{\mbox{CaPtain-Harness}}
\newcommand{\eg}{\emph{e.g.}\xspace}

\makeatletter
\def\newlabel#1#2{\@newl@bel r{#1}{#2}\@newl@bel r{S-#1}{#2}\@newl@bel r{M-#1}{#2}}
\makeatother

\makeatletter
\newcommand{\teaserfigure}{}
\newcommand{\setteaser}[1]{\renewcommand{\teaserfigure}{#1}}
\renewcommand{\mymaketitle}{%
  \tcbset{enhanced,frame hidden,left=0.5cm,right=0.5cm,top=0.45cm,bottom=0.4cm,arc=16pt,before skip=0pt,grow to left by=1.5pt,grow to right by=1.5pt}
  \begin{tcolorbox}
    \setlength{\parindent}{0cm}\setlength{\parskip}{0cm}
    {\centering\nohyphens\titlelist\par\vskip 0.25cm\authorlist\par\vskip 0.2cm\affiliationlist\par\contributionlist\par}
    \vskip 0.3cm
    \teaserfigure
    \ifdefempty{\metadatalist}{}{\vskip 0.2cm\metadatalist\par}
  \end{tcolorbox}
  \tcbset{reset}
  \vskip 0.25cm
  {\centering\large\sffamily\bfseries Abstract\par}\vskip 0.15cm
  {\small\abstractlist\par}
  \vskip 0.2cm
}
\makeatother

\title{\method{}: Post-Training Open Robot-Use Agents with a Physics Critic}
\renewcommand{\titlelist}{{\fontsize{22}{28}\selectfont\sffamily\bfseries \method{}: Post-Training Open\\ Robot-Use Agents with a Physics Critic\par}}

\author{Yijia~Fan}
\author{Xilin~He}
\author{Yilei~Jiang}
\author{Dongming~Wu}
\author{Jiaxu~Wang}
\author[\dagger]{Xiangyu~Yue}
\affiliation{MMLab, The Chinese University of Hong Kong \qquad NovaXBot}
\contribution[\dagger]{Corresponding author: \href{mailto:xyyue@ie.cuhk.edu.hk}{\texttt{xyyue@ie.cuhk.edu.hk}}}

\abstract{\input{sections/abstract.tex}}

\makeatletter\renewcommand\metadataformat[2][]{{\centering\small\sffamily\bfseries #1\par}}\makeatother
\newcommand{\linkicon}[1]{\raisebox{-0.18em}{\includegraphics[height=0.95em]{figs/icons/#1}}}
\metadata[\linkicon{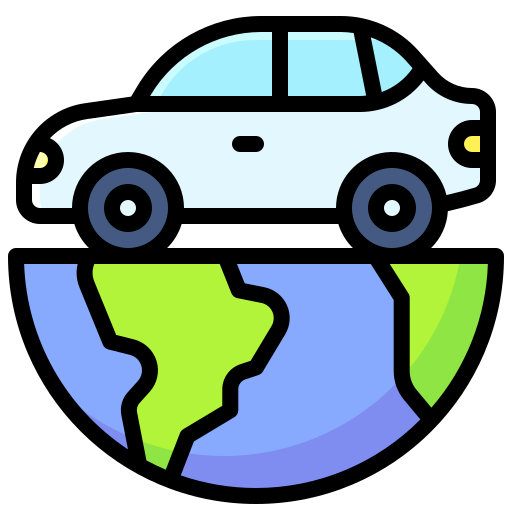}~\href{https://waltstephen.github.io/captain/}{Project Page}\quad$\cdot$\quad\linkicon{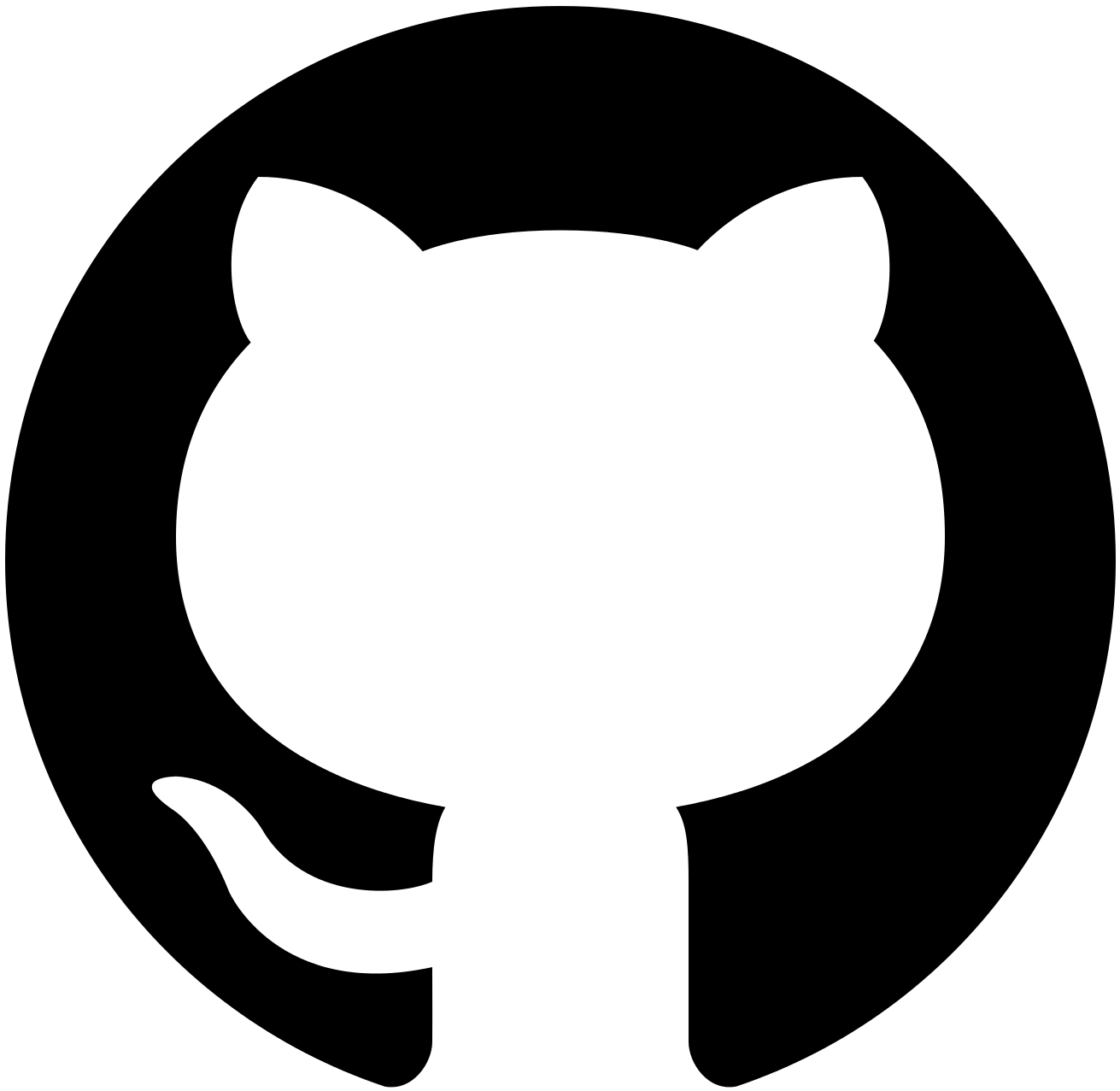}~\href{https://github.com/waltstephen/CaPtain}{Code}\quad$\cdot$\quad\linkicon{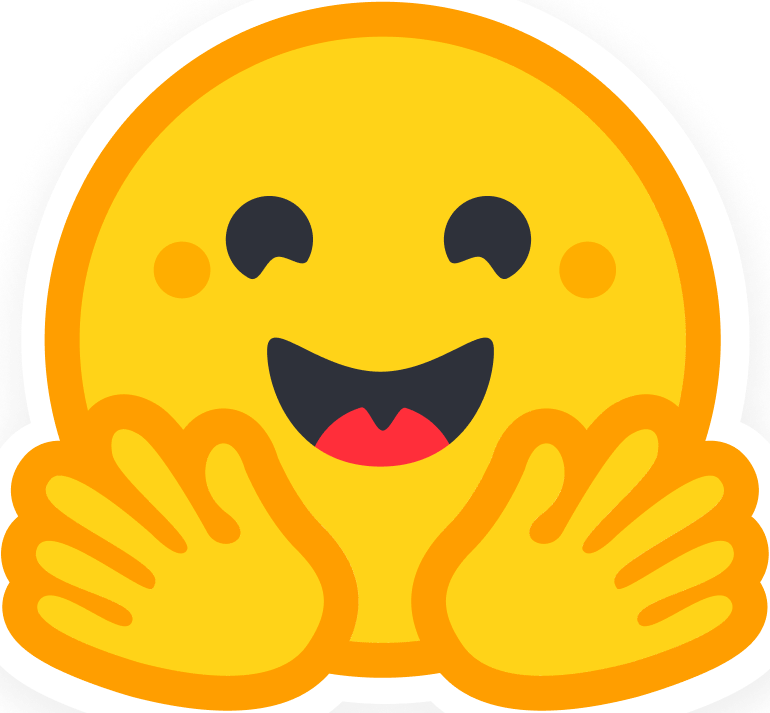}~\href{https://huggingface.co/collections/YijiaFan/captain-6ae3314bcf088fae95904168}{Models \& Data}]{}

\setteaser{%
  \centering\includegraphics[width=\linewidth]{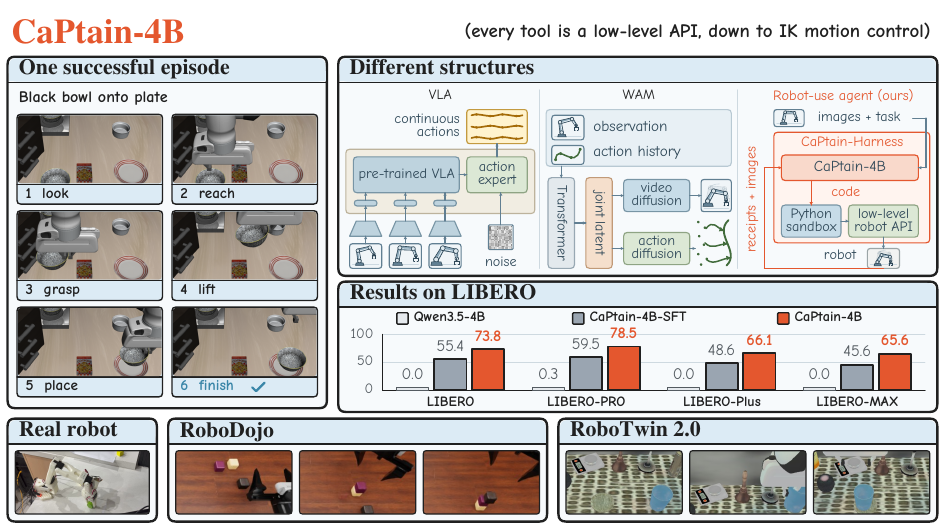}\par\vskip 0.1cm
  \captionof{figure}{\textbf{\model{}} is a robot-use agent: a 4B open model, trained with \textbf{\method{}}, that controls a robot by writing code against low-level primitives. RL with our \critic{} improves it over SFT across LIBERO and three perturbation suites; it reaches 78.5 on LIBERO-PRO and transfers to new benchmarks without training on them.}\label{fig:teaser-front}}
\hypersetup{citecolor=black}
\begin{document}
\raggedbottom
\vspace*{-1.1cm}
\maketitle

\input{sections/intro.tex}
\input{sections/related.tex}
\input{sections/method.tex}
\input{sections/experiments.tex}
\input{sections/rest.tex}

\bibliography{refs}
\bibliographystyle{plainnat}

\clearpage
\appendix
\section*{Appendix}
\vspace{-0.2cm}
{
\small
\setlength{\parskip}{1pt}
\setlength{\baselineskip}{1.18\baselineskip}
\startcontents[appendices]
\printcontents[appendices]{l}{1}{\setcounter{tocdepth}{2}}
}
\clearpage
\FloatBarrier
\input{sections/appendix.tex}

\end{document}

%% file: sections/abstract.tex
Robot-use agents operate a robot the way computer-use agents operate a computer: they observe camera images, write code against perception and motion primitives, and read back the results.
Current capable agents are not trained: they prompt closed frontier models in every episode or delegate manipulation to vision-language-action (VLA) skills, while small open vision-language models (VLMs) fail outright.
We present \method{}, a post-training framework that turns a small open VLM into a robot-use agent that makes every manipulation decision itself.
\harness{} exposes only low-level perception, measurement, and guarded-motion primitives, returns a text receipt after every call, and is shared across five simulated benchmarks and a real robot.
In-harness supervised fine-tuning (SFT) teaches the interface, and multi-turn reinforcement learning (RL) refines the agent from its own interactions.
To credit tens of requests from a single outcome, RL uses a \critic{} that reads privileged simulator state, such as object poses and grasp states, and credits each request by its change in predicted success; the critic is removed at test time.
From Qwen3.5-4B (0\% zero-shot), \model{} reaches 73.8\% on held-out LIBERO (55.4\% after SFT) and 78.5\% on LIBERO-PRO, the best open-weight agent without a VLA skill, above several frontier-model agents.
Its RL gains hold on three perturbation suites (+17.6 to +21.2 points), and removing critic credit costs 7.1 points.
Without training on the target benchmarks, it reaches 83.3\% on SimplerEnv WidowX and beats Gemini-3-Pro on CaP-Bench; adapted only in simulated twins, it runs on two real arms without real-world data.

%% file: sections/intro.tex
\section{Introduction}
\label{sec:intro}

Computer-use and coding agents operate a computer by issuing commands and reading back the results, and they are now routinely post-trained with supervised fine-tuning (SFT) followed by multi-turn reinforcement learning (RL)~\citep{katcoder2026,uitars2,sao2026,feng2025gigpo}.
A \emph{robot-use agent} operates a robot in the same way: it observes camera images, writes code against perception and motion primitives, and reads back what physically happened.
It builds on language models that write robot programs~\citep{liang2023cap,huang2023voxposer,qin2026agenticrobots}.
Because the program is composed at test time, the agent inherits the generalization of the model that writes it. Frontier models make such agents strikingly capable: used directly as a robot policy, GPT-6 Astra ranks above every public entry on the RoboDojo leaderboard~\citep{astra2026robodojo}, and with a coding harness it solves 94.0\% of LIBERO-PRO episodes~\citep{pyrualean2026}.

Existing robot-use agents, however, are not trained.
One line of work prompts closed frontier models in every episode~\citep{openrua2026,racap2026,waa2026,aspire2026,k1harness2026}; these agents cannot be adapted to new robots or tasks and cannot run on onboard compute.
Because a robot-use agent acts through tens of short calls per episode, the latency of each call directly sets how long the robot waits between motions and how quickly it reacts when the scene changes.
Another line of work delegates manipulation to pre-trained vision-language-action models (VLAs) or skills~\citep{pyrualean2026,agenticroborl2026,sensenovaroborsi2026,argon2026robotics}: the code decides \emph{which} skill to call, but the skill decides \emph{how} to grasp, approach, and place, so learning can at best re-sequence skills.
Small open vision-language models (VLMs) fail outright when asked to control a robot directly; Qwen3.5-4B solves none of the LIBERO episodes in our harness.
This raises a natural question: \emph{can a small open VLM be post-trained, like a computer-use agent, into a robot-use agent that makes every manipulation decision itself?}

Post-training at the level of low-level primitives requires three things.
First, an \emph{interface} that exposes every manipulation decision to learning: where to grasp, from which direction to approach, and when to close the gripper must be written by the policy, not hidden inside a skill.
Second, a \emph{starting point}: a base model that never completes an episode cannot explore productively in RL.
Third, and most critically, \emph{credit assignment}: an episode spans tens of requests, most of which only perceive or verify, while the outcome is observed once at the end and physical effects often surface several turns later.
Robot learning offers a lever here that language agents lack: training runs in simulation, where object poses, grasp states, and sub-goal progress are known exactly, while the deployed policy acts from its cameras alone.

We present \textbf{\method{}}, a post-training framework that turns a small open VLM into a robot-use agent, with one component for each requirement.
\textbf{\harness{}} exposes 22 low-level perception, measurement, and guarded-motion primitives and nothing higher, returns a text receipt of what physically happened after every call, and keeps the same policy-facing API across five simulated benchmarks and a real robot.
\textbf{In-harness SFT} on synthetic trajectories in the exact multi-turn format of deployment teaches the base model the interface.
\textbf{Multi-turn RL with a \critic{}} then refines the agent from its own interactions, building on streaming RL~\citep{sao2026} and asymmetric actor-critic training~\citep{pinto2017asymmetric}. The critic reads the simulator's privileged physical state and credits each request by how much it changes the predicted probability of success, evaluated at a fixed request index so that elapsed time alone earns no credit; at test time, the policy sees only images and receipts.

Applied to Qwen3.5-4B, \method{} yields \textbf{\model{}}.
It raises held-out LIBERO success from 0\% to 55.4\% after SFT and to 73.8\% after RL. On LIBERO-PRO, it reaches 78.5\%, ranking highest among open-weight agents without a VLA skill and outperforming the frontier-model agents K1~\citep{k1harness2026} (77.2\%), WAA~\citep{waa2026} (75.6\%), and ASPIRE~\citep{aspire2026} (71.7\%).
In the same harness it also exceeds GPT-6 Astra (68.5\%), the frontier model that generated most of its SFT data; after SFT alone it is below (59.5\%).
The RL gains hold under perturbation, from +17.6 to +21.2 points across LIBERO-PRO~\citep{liberopro2025}, LIBERO-Plus~\citep{liberoplus2025}, and LIBERO-MAX~\citep{liberomax2026}.
The \critic{} drives these gains: removing its per-request credit costs 7.1 points, and a critic-free GRPO baseline ends below its SFT initialization.
Without target-domain training, \model{} reaches 83.3\% on SimplerEnv WidowX and 75.0 on CaP-Bench, surpassing Gemini-3-Pro, and completes tasks on a real robot after adaptation in simulation alone.
Running locally on two consumer GPUs at 1.2\,s per request, it completes tasks 4--15$\times$ faster than frontier models in the Codex harness and re-grasps a moved object within seconds where they take minutes.

\paragraph{Contributions.}
\begin{enumerate}
    \item \textbf{\method{}}, a post-training framework for robot-use agents: in-harness SFT followed by multi-turn RL with a \critic{} that turns privileged physical state into per-request credit and is removed at test time.
    \item \textbf{\harness{}}, a low-level robot-use interface with grounded text receipts and without skills or a VLA, shared unchanged across five simulated benchmarks and a real robot.
    \item \textbf{Open \model{}, data, and infrastructure}: model weights, in-harness SFT data, and the full training infrastructure (harness, SFT code, and streaming multi-turn RL system).
\end{enumerate}

%% file: sections/related.tex
\section{Related Work}
\label{sec:related}

\paragraph{Robot-use agents.}
Code as Policies~\citep{liang2023cap} and VoxPoser~\citep{huang2023voxposer} showed that a language model can control a robot by writing programs over perception and control APIs.
Recent agents scale this with frontier models and richer harnesses~\citep{astra2026robodojo,argon2026robotics,architect2026,aor2026}.
Many reach their headline results with frontier models and no weight updates~\citep{aspire2026,rats2026,rho2026,racap2026,openrua2026,robodawn2026,encore2026,roborsi2026}, at up to \$22 per episode~\citep{openrua2026}.
Others orchestrate a frozen VLA as a skill~\citep{harnessvla2026,pyrualean2026,pigey2026,agenticroborl2026,agenticrobot2025,adahvla2026} or use privileged simulator state or a success oracle at test time~\citep{faea2026,k1harness2026,waa2026}.
Few works train the code-writing model.
CaP-X~\citep{fu2026capx} applies GRPO to a single-turn, text-only 7B code model, while Guava~\citep{guava2026} trains Qwen3.5-4B primarily via SFT.
PyRUA~\citep{pyrualean2026}, SenseNova-RoboRSI~\citep{sensenovaroborsi2026}, and AgenticRoboRL~\citep{agenticroborl2026} train agents on top of VLA, motion-planner, or skill primitives.
\method{} trains a small open VLM directly over low-level primitives to act from images and receipts alone.

\paragraph{End-to-end policies and perturbation suites.}
VLAs~\citep{kim2024openvla,kim2025oft,black2024pi0,pi05,gr00tn1,gr00tn15,liu2024rdt} and world-action models~\citep{openwam2026pretrain,openwam2026framework,robustwam2026,mowam2026} reach 94--98\% on standard LIBERO, and RL post-training of VLAs raises in-distribution success~\citep{simplevlarl,riptvla,rl4vla,pirl,vlarl,rlinfvla,sop2026}.
Perturbation suites~\citep{liberopro2025,liberoplus2025,liberomax2026} show that these policies remain sensitive to changes in objects, layouts, and instructions, even after RL post-training~\citep{riptvla}.
We find that the RL gains of a code-writing policy persist across all three suites (\S\ref{sec:robust}).

\paragraph{Credit assignment and privileged critics.}
PPO~\citep{schulman2017ppo} with GAE~\citep{schulman2015gae} and critic-free GRPO~\citep{shao2024deepseekmath} are standard for LLM RL, and turn-level credit assignment for multi-turn agents is an active topic~\citep{turnppo,capo2026,vldac,era2025,hygae2026,vagen2025,zhou2024archer,feng2025gigpo,spo2025,sapo2026,creditsurvey2026,kazemnejad2024vineppo,sao2026}.
The closest works convert a value or progress difference into turn-level credit using a game solver~\citep{cast2026}, acceptance checks~\citep{procredit2026}, a distance estimator~\citep{t2spo2026}, or rule-based simulator predicates~\citep{dls2026}.
Privileged critics for LLMs~\citep{sweetrl2025,lecritique2026,katcoder2026,odysseus2026} do not observe physical state.
In robotics, asymmetric actor-critic~\citep{pinto2017asymmetric,baisero2021unbiased,informedaac2025} and privileged teachers~\citep{chen2019lbc,lee2020quadruped,kumar2021rma} use full simulator state for non-language actors, including on LIBERO~\citep{mtlibero,agenticexec2026}.
\method{} brings this asymmetry to a code-writing language-model actor and learns a value function over privileged physical state, since continuous manipulation has neither a solver nor a checklist.

%% file: sections/method.tex
\section{\method{}}
\label{sec:method}

\begin{figure*}[t]
\centering
\includegraphics[width=\linewidth]{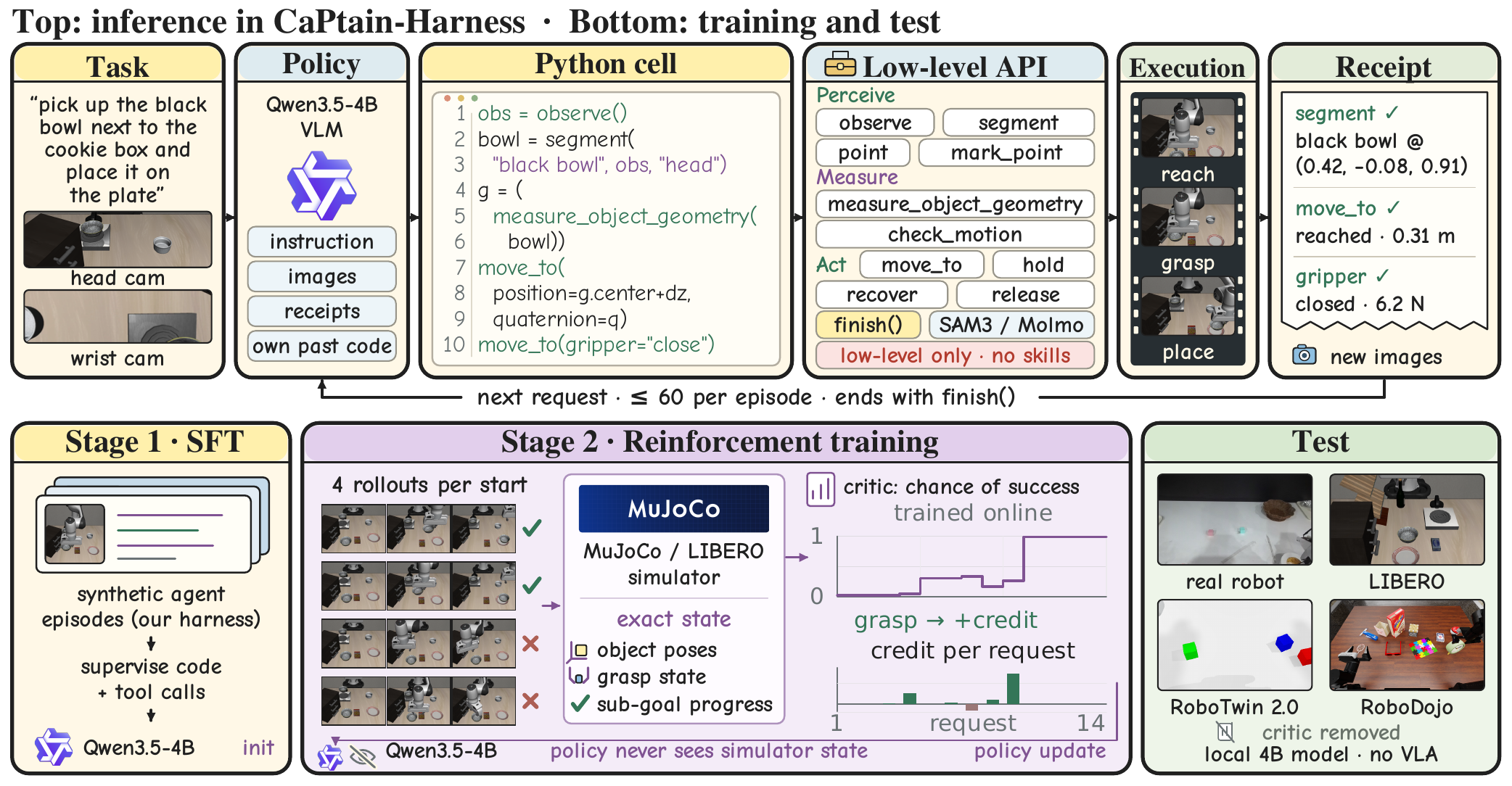}
\caption{\textbf{Overview of \method{}.}
\emph{Top:} inference, one episode in \harness{}: at each request the policy writes a Python cell against a low-level API and receives text receipts and new images, for up to 60 requests. \texttt{measure\_object\_geometry} returns measurements only (center, extents, heights); the grasp offset \texttt{dz} and orientation \texttt{q} are computed by the policy.
\emph{Bottom:} training and test: SFT on synthetic agent trajectories, then RL in which a \critic{} turns privileged physical state into per-request credit; the critic is removed at test time.}
\label{fig:method}
\end{figure*}

\method{} trains a small open VLM as a robot-use agent (Figure~\ref{fig:method}). The agent acts in \harness{} (\S\ref{sec:harness}), is initialized by SFT in the harness (\S\ref{sec:sft}), and is trained with multi-turn RL in which a \critic{} credits each request (\S\ref{sec:rl}). Implementation details are in Appendices~\ref{S-app:impl} and~\ref{S-app:infra}--\ref{S-app:lessons}.

\subsection{\harness{}}
\label{sec:harness}

\paragraph{A ReAct loop over requests.}
A task is given as a natural-language instruction and a \texttt{TASK.md} file that describes the API, the coordinate frames and the success criteria.
The agent runs a ReAct-style loop~\citep{yao2023react,argus2026} in which actions are code.
An episode is a sequence of \emph{requests} $t=1,\dots,T$ with $T\le 60$.
At request $t$ the policy observes the history $h_t=(\ell,o_1,a_1,\dots,a_{t-1},o_t)$, where $\ell$ is the task description with the initial head and wrist images, $a_k$ is the output of request $k$, and $o_k$ is what the harness returned.
Each request is one model call, $a_t\sim\pi_\theta(\cdot\mid h_t)$: reasoning followed by one or more calls to a persistent \texttt{python} tool, whose cell results, receipts and images form $o_{t+1}$.
The loop ends when a reply contains no tool call, normally after \texttt{finish()}, or when a budget on requests, physical actions or wall time runs out.
Context management, the Python session and backend portability are in Appendix~\ref{S-app:harness-details} (overview in Figure~\ref{S-fig:harness}).

\paragraph{Low-level primitives only.}
The \texttt{robot} module exposes 22 capabilities plus \texttt{finish()} (Appendix~\ref{S-app:api}).
\emph{Perception} includes RGB-D observation, text-prompted segmentation with SAM~3~\citep{sam3}, pointing with Molmo2~\citep{clark2026molmo2}, and \texttt{mark\_point}, which back-projects a pixel the model chose.
\emph{Measurement} returns object geometry and contact checks.
\emph{Motion} consists of \texttt{move\_to} (an end-effector pose or a jaw command), \texttt{hold}, \texttt{recover} and \texttt{release}; read-only planning and state queries complete the API.
There are no grasp, place or open-drawer skills and no VLA.

\paragraph{Division of labor.}
The harness \emph{measures}, \emph{checks} and \emph{executes}; the policy \emph{decides}.
Measurement tools return what the cameras show: masks and metric points of detections, and object geometry (center, extents, top and bottom heights, footprint, yaw and the offset to a destination); they never return a grasp pose or an approach.
Check tools such as \texttt{check\_motion} and \texttt{inspect\_contact\_geometry} only evaluate a pose that the policy has already proposed, reporting reachability or how the jaws would meet the object, without executing anything.
Execution turns a commanded end-effector pose into a straight Cartesian motion with inverse kinematics, self-collision checks and force monitoring, and a jaw command into a gripper motion.
Everything in between is computed by the policy in its own code: the grasp position derived from the measured geometry, the gripper orientation, the pre-grasp and approach heights, the speed and contact mode, when to close and open the jaw, where to release, and when to call \texttt{finish()} (Figure~\ref{fig:method}; full episodes in Appendix~\ref{S-app:example}).
Every approach vector, grasp pose and jaw command is therefore exposed to post-training.

\paragraph{Receipts and success.}
After each robot call, the harness appends a short \emph{receipt} of what physically happened: for motions, the stop reason, executed distance, peak force and jaw opening before and after; for perception, the detection identity, metric position and offset to the destination.
The agent can also attach up to four camera images per cell.
A completion gate checks the agent's recorded measurements and may reject \texttt{finish()} twice before accepting it.
\emph{Host success} is the benchmark's goal predicate on the final state; \emph{strict success} also requires the agent to end the episode with \texttt{finish()}, since a robot that does not declare completion cannot be deployed.

\subsection{Initialization: SFT in the harness}
\label{sec:sft}

Qwen3.5-4B~\citep{qwen35blog} scores 0\% zero-shot in \harness{}, so we first fine-tune it on synthetic agent trajectories that frontier coding agents, mainly GPT-6 Astra, generated in the harness: about 26k examples and 0.86B tokens, a third of them visual QA (Appendix~\ref{S-app:sft}).
Trajectories use the same chat template, system prompt, \texttt{TASK.md}, tool schema and receipt format as in RL and at test time.
Given $\tau=(h_1,a_1,\dots,h_T,a_T)$, we supervise only agent tokens:
\begin{equation}
\mathcal{L}_{\mathrm{SFT}}(\theta)=-\sum_{t=1}^{T}w_t\sum_{j=1}^{|a_t|}\log\pi_\theta\big(a_{t,j}\mid h_t,a_{t,<j}\big),
\end{equation}
where $w_t=0$ masks turns with repeated execution failures or recovery loops.
Visual question-answering examples are added with weight $0.5$.
Since $a_t$ includes reasoning, the model learns to plan and read receipts, not only to emit code.
To prevent the model from copying pixel coordinates from earlier \texttt{mark\_point} calls, we replace them in the history with a mask token with probability $0.2$.
We train all language-model weights and freeze the vision tower.

\subsection{RL with the \critic{}}
\label{sec:rl}

RL maximizes $J(\theta)=\mathbb{E}_{\tau\sim\pi_\theta}\big[R(\tau)+\sum_t c_t\big]$, where $R$ is a terminal return and $c_t\le0$ are small per-request costs.
With low-level primitives, an episode has dozens of requests and one sparse outcome, and each request must be assigned credit for its share of that outcome.

\paragraph{Privileged state.}
Before each request, the simulator records a snapshot $s_t$ that the policy never sees.
From it we compute 36 features $x(s_t)$: end-effector and jaw state; goal bookkeeping; distances between end-effector, target and destination; target lift and displacement; and articulated-joint positions with held flags (Appendix~\ref{S-app:impl}).
We exclude counters of the policy's own behavior, such as the number of perception calls or failed calls; they act as proxies for elapsed time and absorb credit from grasping (Appendix~\ref{S-app:criticdesign}).

\paragraph{The \critic{}.}
The critic predicts host success from the physical state, the request index and the task:
\begin{equation}
V_\phi(s_t,t)=\sigma\Big(\tfrac{1}{5}\textstyle\sum_{k=1}^{5} f_{\phi_k}\big([x(s_t);\,t;\,e_{\text{task}}]\big)\Big),
\end{equation}
an ensemble of five small MLPs with averaged logits.
It is warm-started on episodes of an earlier policy and co-trained online as in SAO~\citep{sao2026}, with a few steps per round on
\begin{equation}
\begin{split}
\mathcal{L}_V(\phi)=-\sum_{i}\sum_{t}\Big[&y_i\log V_\phi(s_{i,t},t)\\
&+(1-y_i)\log\big(1-V_\phi(s_{i,t},t)\big)\Big],
\end{split}
\end{equation}
where $y_i\in\{0,1\}$ is the host success of episode $i$.
Each round's advantages use the critic from before that round's update.
The critic is used only in training and is removed at test time.

\paragraph{Same-$t$ credit.}
The credit of request $t$ is the change in predicted success it causes, plus its cost:
\begin{equation}
\begin{aligned}
\delta_t &= c_t + V_\phi(s_{t+1},t) - V_\phi(s_t,t)\quad(t<T),\\
\delta_T &= c_T + R_i - V_\phi(s_T,T).
\end{aligned}
\label{eq:td}
\end{equation}
Both terms of each difference use the same index $t$.
Predicted success falls as the request budget is used up, so a standard TD error with $V(s_{t+1},t{+}1)$ would charge every request for the passage of time.
Holding $t$ fixed removes this term: the value difference in $\delta_t$ is nonzero only if the physical state changed.
For a request that only perceives or verifies, the one-step $\delta_t$ is therefore near zero, however many tokens it contains.
Such requests still receive credit through generalized advantage estimation~\citep{schulman2015gae}, $A^{\mathrm{GAE}}_t=\sum_{k\ge0}\lambda^k\delta_{t+k}$ with $\lambda=0.9$, from the physical actions they enable.
Unlike standard TD errors, these $\delta_t$ do not telescope to $R_i-V_\phi(s_1,1)$: they measure physical progress, not the return, which enters through a group-relative outcome term over $G{=}4$ rollouts per start state:
\begin{equation}
\begin{split}
A_{i,t} = {}&\big(G_i - \overline{G}_{\text{group}}\big)\\
&+ \big(A^{\mathrm{GAE}}_{i,t} - \overline{A^{\mathrm{GAE}}_{i,\cdot}}\big),
\end{split}
\label{eq:adv}
\end{equation}
where $G_i = R_i + \textstyle\sum_t c_{i,t}$.
The first term ranks rollouts from the same start state; the second ranks requests within a rollout.
Advantages are divided by a running standard deviation, and all tokens of request $t$, including its reasoning, share $A_{i,t}$.
A request whose cells produced only Python errors receives no positive credit and incurs a small extra penalty.

\paragraph{Returns and costs.}
The terminal return separates outcomes that matter for deployment: $R_i=1$ for success with \texttt{finish()}, $0.25$ for success without it, $0$ for other failures, $-0.1$ when the budget is exhausted, and $-0.3$ for \texttt{finish()} on a failed state.
The costs $c_t$ penalize failed or rejected physical actions, physical actions well beyond the typical count for the task, and any activity after the scene is solved.
If all goals have held for 15 requests without \texttt{finish()}, the episode is stopped as a missed finish.
On the hardest long-horizon tasks, failed episodes receive small one-shot sub-goal rewards, bounded below a missed finish (Appendix~\ref{S-app:reward}).

\paragraph{Policy update.}
We use the streaming single-pass update of SAO~\citep{sao2026} with double-sided token masking.
With $r_{t,j}=\pi_\theta(a_{t,j}\mid\cdot)/\pi_{\text{old}}(a_{t,j}\mid\cdot)$,
\begin{equation}
\begin{split}
\mathcal{L}_\pi(\theta)={}&-\textstyle\sum_{i,t,j} m_{t,j}\, r_{t,j}\, A_{i,t}\\
&+\beta\,\mathrm{KL}_{k3}\big(\pi_\theta\,\|\,\pi_{\mathrm{SFT}}\big),
\end{split}
\end{equation}
with $m_{t,j}=\mathbf{1}[0.2<r_{t,j}<4]$.
Tokens whose ratio drifts too far are dropped rather than clipped; this also bounds the effect of asynchronous rollouts up to two policy versions old.
At most eight requests per trajectory enter the policy gradient, sampled uniformly and reweighted by inclusion probability; the critic uses every request.
The gradient-noise scale is in the thousands of episodes (Appendix~\ref{S-app:noise}), so we accumulate over 1{,}024 episodes per optimizer step (pseudocode in Appendix~\ref{S-app:rlround}).

\begin{figure*}[t]
\centering
\includegraphics[width=\linewidth]{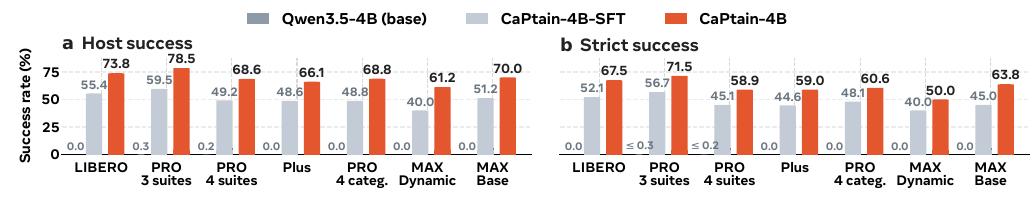}
\caption{\textbf{Base model, SFT, and RL across the LIBERO benchmark family}: (a) host and (b) strict success (\%) on the same episodes. Strict success of the base model is bounded by its host success ($\leq$). Episode counts and exact McNemar tests ($p<10^{-5}$ on all primary suites) are in Appendix~\ref{S-app:results} and Table~\ref{S-tab:main}.}
\label{fig:main}
\end{figure*}

\begin{table*}[t]
\centering\small
\setlength{\tabcolsep}{4pt}
\caption{\textbf{Comparison with published systems on LIBERO-PRO} (success \%).
\emph{Frontier API}: queries a closed frontier model at test time; \emph{VLA Skill}: invokes a policy trained on LIBERO demonstrations; \emph{In-Domain Data}: uses LIBERO demonstrations or LIBERO-PRO development states; \emph{In \harness{}}: our runs on the same 600 three-suite episodes.
Protocol notes are in Appendix~\ref{S-app:results}; the full LIBERO-family comparison is in Appendix~\ref{S-app:fullcompare}.}
\label{tab:baselines}
\begin{tabular*}{\linewidth}{@{\extracolsep{\fill}}ll ccc cc@{}}
\toprule
 & & Frontier & VLA & In-Domain & \multicolumn{2}{c}{LIBERO-PRO (\%)} \\
\cmidrule(l){6-7}
System & Policy Model & API & Skill & Data & 3 Suites & All 4 Suites \\
\midrule
\multicolumn{7}{@{}l}{\textit{End-to-end VLAs \& VLA-assisted agents}} \\
Harness VLA~\citep{harnessvla2026} & Claude Opus 4.7 + $\pi_{0.5}$ & \ding{51} & \ding{51} & \ding{51} & 87.7 & 82.4 \\
AgenticRoboRL~\citep{agenticroborl2026} & Qwen3.5-9B + RL-tuned $\pi_{0.5}$ & \ding{55} & \ding{51} & \ding{51} & 81.3 & 71.0 \\
RLinf $\pi_{0.5}$~\citep{harnessvla2026} & $\pi_{0.5}$ (RL fine-tuned) & \ding{55} & \ding{51} & \ding{51} & 56.2 & 50.0 \\
$\pi_{0.5}$~\citep{liberopro2025} & $\pi_{0.5}$ (SFT) & \ding{55} & \ding{51} & \ding{51} & 12.8 & 10.8 \\
\midrule
\multicolumn{7}{@{}l}{\textit{Closed frontier-model robot-use agents}} \\
OpenRUA~\citep{openrua2026} & Claude Opus 5 & \ding{51} & \ding{55} & \ding{55} & 91.8 & 87.0 \\
K1~\citep{k1harness2026} & Gemini 3.7 Flash & \ding{51} & \ding{55} & \ding{51} & 77.2 & -- \\
WAA~\citep{waa2026} & Gemini 3.7 Flash & \ding{51} & \ding{55} & \ding{55} & 75.6 & -- \\
ASPIRE~\citep{aspire2026} & Claude Opus 4.6 & \ding{51} & \ding{55} & \ding{51} & 71.7 & -- \\
RACaP~\citep{racap2026} & GPT-5.5 & \ding{51} & \ding{55} & \ding{51} & 45.0 & -- \\
RATs~\citep{rats2026} & Gemini 3.1 Pro & \ding{51} & \ding{55} & \ding{51} & 43.8 & -- \\
CaP-Agent0~\citep{fu2026capx} & 3-model ensemble & \ding{51} & \ding{55} & \ding{55} & 18.2 & -- \\
\midrule
\multicolumn{7}{@{}l}{\textit{Open-weight robot-use agents (no VLA skill)}} \\
SPARK~\citep{spark2026} & Qwen3.8-27B & \ding{55} & \ding{55} & \ding{51} & 66.8 & -- \\
Guava~\citep{guava2026} & Qwen3.5-4B (SFT+GRPO) & \ding{55} & \ding{55} & \ding{55} & $\approx$41.0 & -- \\
\midrule
\multicolumn{7}{@{}l}{\textit{In \harness{}, same 600 episodes}} \\
GPT-6 Astra & GPT-6 Astra & \ding{51} & \ding{55} & \ding{55} & 68.5 & -- \\
Guava~\citep{guava2026} & Qwen3.5-4B (SFT+GRPO) & \ding{55} & \ding{55} & \ding{55} & 13.8 & -- \\
\rowcolor{oursbg!45} \model{}-SFT (ours) & Qwen3.5-4B (SFT) & \ding{55} & \ding{55} & \ding{55} & 59.5 & 49.2 \\
\rowcolor{oursbg} \textbf{\model{} (ours)} & \textbf{Qwen3.5-4B (SFT+RL)} & \ding{55} & \ding{55} & \ding{55} & \textbf{78.5} & \textbf{68.6} \\
\bottomrule
\end{tabular*}
\end{table*}

\begin{table*}[t]
\centering\footnotesize
\setlength{\tabcolsep}{2.5pt}
\caption{\textbf{Zero-shot transfer.}
Trained only on LIBERO, \model{} is evaluated without further training against VLAs/WAMs trained on target-benchmark demonstrations and robot-use agents (RUA; success \%).
CaP-X (M/S): multi-/single-turn; CaP-RL is trained with RL on CaP-Bench; RoboDojo in Appendix~\ref{S-app:robodojo}.}
\label{tab:transfer}
\begin{minipage}[t]{0.33\linewidth}
\centering
\textbf{(a) RoboTwin~2.0} (Hard, 5{,}000 episodes)\par\smallskip
\begin{tabular*}{\linewidth}{@{\extracolsep{\fill}}llrr@{}}
\toprule
Method & Type & All & 1-Arm \\
\midrule
PyRUA-Lean-4B~\citep{pyrualean2026} & RUA+VLA & 52.4 & -- \\
OpenWAM-$\alpha$~\citep{openwam2026pretrain} & WAM & 48.7 & -- \\
$\pi_{0.5}$~\citep{pi05} & VLA & 46.0 & 45.8 \\
X-VLA~\citep{xvla2025} & VLA & 20.9 & -- \\
Xiaomi-R0~\citep{xiaomir02026} & VLA & 18.2 & -- \\
$\pi_0$~\citep{black2024pi0} & VLA & 16.3 & 18.3 \\
RDT~\citep{liu2024rdt} & VLA & 13.7 & 14.0 \\
\rowcolor{oursbg} \textbf{\model{}} & \textbf{RUA} & \textbf{22.6} & \textbf{30.0} \\
\bottomrule
\end{tabular*}
\end{minipage}\hfill
\begin{minipage}[t]{0.3\linewidth}
\centering
\textbf{(b) SimplerEnv} (WidowX Visual Matching)\par\smallskip
\begin{tabular*}{\linewidth}{@{\extracolsep{\fill}}llr@{}}
\toprule
Method & Backbone & SR \\
\midrule
ManiAgent~\citep{maniagent2025} & GPT-5 & 86.8 \\
ManiAgent~\citep{maniagent2025} & Grok-4 & 83.0 \\
PhysBrain~1.0~\citep{physbrain2026} & VLA & 80.2 \\
SoFar~\citep{sofar2025} & Agent & 58.3 \\
$\pi_{0.5}$~\citep{pi05} & VLA & 57.1 \\
$\pi_0$~\citep{black2024pi0} & VLA & 27.1 \\
OpenVLA-OFT~\citep{kim2025oft} & VLA & 23.4 \\
\rowcolor{oursbg} \textbf{\model{}} & \textbf{Qwen3.5-4B} & \textbf{83.3} \\
\bottomrule
\end{tabular*}
\end{minipage}\hfill
\begin{minipage}[t]{0.3\linewidth}
\centering
\textbf{(c) CaP-Bench} (Cube Lift \& Stack)\par\smallskip
\begin{tabular*}{\linewidth}{@{\extracolsep{\fill}}llr@{}}
\toprule
Method & Backbone & SR \\
\midrule
CaP-X (M)~\citep{fu2026capx} & Gemini-3-Pro & 73.5 \\
CaP-RL~\citep{fu2026capx} & Qwen2.5-7B & 62.0 \\
CaP-X (M)~\citep{fu2026capx} & GPT-5.2 & 55.5 \\
CaP-X (M)~\citep{fu2026capx} & Opus 4.5 & 54.0 \\
CaP-X (S)~\citep{fu2026capx} & Qwen2.5-7B & 14.5 \\
Base model & Qwen3.5-4B & 5.0 \\
\rowcolor{oursbg!45} \model{}-SFT & Qwen3.5-4B & 55.0 \\
\rowcolor{oursbg} \textbf{\model{}} & \textbf{Qwen3.5-4B} & \textbf{75.0} \\
\bottomrule
\end{tabular*}
\end{minipage}
\end{table*}
\begin{table}[t]
\centering\footnotesize
\setlength{\tabcolsep}{3pt}
\caption{\textbf{RL ablations} on held-out LIBERO (240 episodes). Each variant is one 128-round RL run (32 optimizer updates) from \model{}-SFT that changes one component. Successes as episode counts (\%); $\Delta$ in points vs.\ \model{} at round 128.}
\label{tab:ablation}
\begin{tabular*}{\linewidth}{@{\extracolsep{\fill}}lcccc@{}}
\toprule
Variant & Host & Strict & $\Delta$H & $\Delta$S \\
\midrule
\rowcolor{oursbg} \model{} (round 128) & 174 (72.5) & 159 (66.2) & -- & -- \\
\midrule
w/o critic credit & 157 (65.4) & 142 (59.2) & $-7.1$ & $-7.1$ \\
w/o same-$t$ & 167 (69.6) & 134 (55.8) & $-2.9$ & $-10.4$ \\
w/o early stop & 151 (62.9) & 122 (50.8) & $-9.6$ & $-15.4$ \\
policy value head (MC) & 0 (0.0) & 0 (0.0) & $-72.5$ & $-66.2$ \\
critic-free (GRPO) & 119 (49.6) & 82 (34.2) & $-22.9$ & $-32.1$ \\
\midrule
\model{}-SFT (no RL) & 133 (55.4) & 125 (52.1) & $-17.1$ & $-14.2$ \\
\bottomrule
\end{tabular*}
\end{table}

\paragraph{Training distribution.}
RL samples fresh start states across all 40 LIBERO tasks, mostly wide-layout scenes with randomized lighting, camera and materials (Appendix~\ref{S-app:diversity}), and excludes the 240 official evaluation states.
Episodes are stopped early once the critic predicts near-certain failure for several consecutive requests, except in a small control slice.

%% file: sections/experiments.tex
\section{Experiments}
\label{sec:protocol}

\paragraph{Setup.}
We train only on the 40 original LIBERO tasks~\citep{liu2023libero}.
We evaluate on held-out LIBERO initial states 11--16 (240 episodes) and on three perturbation suites never seen in training: LIBERO-PRO~\citep{liberopro2025}, LIBERO-Plus~\citep{liberoplus2025} (stratified 700-case sample) and LIBERO-MAX~\citep{liberomax2026} (80 Lite cases, static and in-episode perturbations); splits and protocols are in Appendix~\ref{S-app:results}.
All evaluations use the RL rollout settings (temperature 1, at most 60 requests per episode, no early stopping) without the \critic{}; object coordinates are estimated online from RGB-D images.
\model{}-SFT is the model after SFT; \model{} is the RL checkpoint at round 300 (75 optimizer updates, 76.8k episodes). RL takes 1.3k H100-hours and SFT 120 (Appendix~\ref{S-app:compute}).

\subsection{Main results}
\label{sec:main}

In Figure~\ref{fig:main}, the base Qwen3.5-4B solves at most 0.3\% on any suite: it exhausts its request budget misusing the interface.
On held-out LIBERO, RL raises host success from 55.4\% to 73.8\% (55 episodes gained, 11 lost) and strict success from 52.1\% to 67.5\%.
False finishes drop from 7.1\% to 2.1\% and requests per episode from 39.1 to 32.6; per suite, host gains range from +13.3 (long) to +26.7 (goal; Appendix~\ref{S-app:expextra}).

\label{sec:robust}
\paragraph{Robustness to perturbations.}
\model{} reaches 78.5\% on LIBERO-PRO (3 suites), 66.1\% on LIBERO-Plus and 61.2\% on LIBERO-MAX Dynamic (Figure~\ref{fig:main}).
The RL gain carries over to every perturbation suite, from +17.6 host points on LIBERO-Plus to +21.2 on LIBERO-MAX Dynamic; on LIBERO-Plus, all seven perturbation dimensions improve at $p<0.05$ and six survive a Bonferroni correction (Appendix Figure~\ref{S-fig:breakdown}b).
Policies trained end-to-end on LIBERO do not hold up: $\pi_{0.5}$ falls from 97\% on LIBERO to 12.8\% on LIBERO-PRO~\citep{liberopro2025}, and the RL-tuned RIPT-VLA stays at its base level on LIBERO-Plus (68.4 vs.\ 69.6~\citep{liberoplus2025}).

\subsection{Comparison with published systems}
\label{sec:compare}

As a standalone 4B policy over low-level primitives, \model{} (78.5\%, Table~\ref{tab:baselines}) is above the open-weight agents SPARK (66.8\%, 27B, ground-truth depth) and Guava ($\approx$41\%, 4B) and the frontier-API agents K1 (77.2\%), WAA (75.6\%) and ASPIRE (71.7\%).
In \harness{} on the same 600 episodes, GPT-6 Astra, which generated most of our SFT trajectories, reaches 68.5\% host and 58.5\% strict success; \model{}-SFT stays below it (59.5\%), and RL takes \model{} 10.0 and 13.0 points beyond it. Guava reaches 13.8\% there: harness and post-training are tightly coupled.
RL in \method{} gains +19.4 points on the 4-suite average, against +5.9 for AgenticRoboRL, which trains a 9B planner around a frozen LIBERO VLA.
A \model{} episode uses 52 GPU-seconds of policy compute (\$0.03--0.06 at \$2--4 per GPU-hour), against \$0.18--22 for published code agents (Appendix~\ref{S-app:cost}).

\subsection{Ablations and analysis of the RL stage}
\label{sec:ablation}

Table~\ref{tab:ablation} removes one component of RL at a time.
\emph{W/o critic credit} keeps the critic for early stopping but gives all requests in an episode the group-relative outcome term of Eq.~\ref{eq:adv}.
\emph{W/o same-$t$} uses the standard difference $V_\phi(s_{t+1},t{+}1)-V_\phi(s_t,t)$.
\emph{W/o early stop} never cuts an episode.
\emph{Policy value head} replaces the \critic{} with a success predictor on the policy's hidden state, using $A_t=R-V(s_t)$ and no early stopping.
\emph{Critic-free (GRPO)} removes critic credit, early stopping and the sub-goal reward, keeping returns, costs, update rule and budgets.
Per-request credit is worth 7.1 points of host and strict success.
Without a fixed request index, strict success drops by 10.4 points and 42.5\% of episodes hit the request cap (vs.\ 28.3\%; Appendix~\ref{S-sec:collapse}).
The critic-free run ends below its SFT initialization: its training success matches \model{} until round 50 and then declines (Appendix~\ref{S-app:ablcurves}); only 82 of 240 test episodes call \texttt{finish()}.
The policy value head collapses: from round 70 the policy ends failing episodes by stopping without \texttt{finish()}, and by round 116 every episode, since $R-V(s_t)$ has no cost terms and penalizes the earliest-quitting failure least (Appendix~\ref{S-app:vbcollapse}).
Without early stopping, the 24--26\% of rollouts judged lost by the critic (Appendix~\ref{S-app:infra}) run to the request cap, adding negative outcome weight to perception and verification requests that successful episodes also need: host and strict success fall by 9.6 and 15.4 points.

\paragraph{What the critic credits.}
On 18.8k episodes sampled across the RL run of \model{} (596k request states), the critic ranks rollouts from the same start state with AUC 0.874 (0.962 overall).
Because Eq.~\ref{eq:td} holds the request index fixed, it rewards requests that change the scene: perception-only requests make up 59\% of requests but receive 0.95\% of the one-step credit mass, whereas grasps account for 4.6\% of requests and receive 31.5\%.
With the standard difference, perception-only requests receive 19.6\%, $20\times$ more, as the time dependence of the value leaks into credit; this explains the strict-success loss in Table~\ref{tab:ablation}.
GAE then passes credit from a physical action back to the perception requests that prepared it (46\% of $|A|$; Appendices~\ref{S-app:criticdesign} and~\ref{S-app:interp}).

\begin{figure*}[t]
\centering
\includegraphics[width=\linewidth]{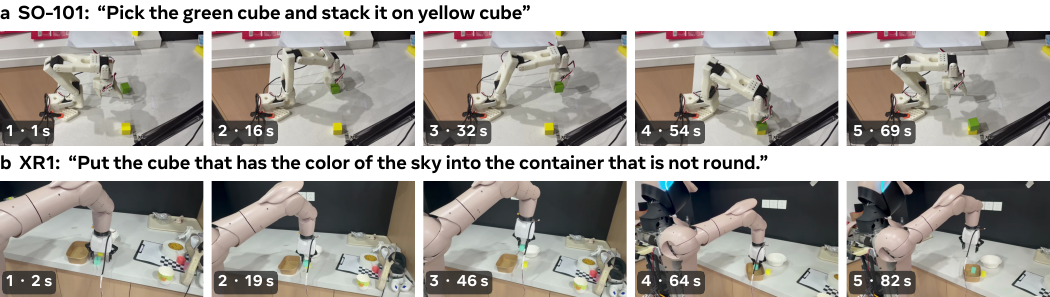}
\caption{\textbf{Real-robot episodes} on (a) the SO-101 and (b) the XR1 with new objects and layouts; five frames each with time stamps (full episodes on the project page). In (b), ``the color of the sky'' is the light-blue cube and ``not round'' the rectangular tray.}
\label{fig:real}
\end{figure*}

\input{sections/transfer.tex}

\subsection{Transfer to real robots}
\label{sec:real}

\begin{table}[t]
\centering\footnotesize
\setlength{\tabcolsep}{4pt}
\caption{\textbf{Real-robot results on the SO-101} with randomized layouts, 10 trials per task, after adaptation in simulation only.}
\label{tab:real}
\begin{tabularx}{\linewidth}{@{}l>{\raggedright\arraybackslash}Xc@{}}
\toprule
Task & Goal & SR (\%) \\
\midrule
Pick & lift the named cube off the table & 100 \\
Pick \& place & put the named cube on the named target & 100 \\
Stack & put one cube on top of another & 80 \\
Ordered placement & put cubes into a bowl in a given color order & 100 \\
\bottomrule
\end{tabularx}
\end{table}

We deploy \model{} on two real arms through hardware backends of \harness{} with an unchanged policy API and no critic, after fine-tuning on episodes from each arm's simulated twin (Appendix~\ref{S-app:realextra}); no real-world data is used.

\paragraph{SO-101.}
The SO-101 is a low-cost five-joint arm with a parallel gripper, a fixed head camera, and a wrist camera; it lacks a force sensor, so its backend estimates contact forces from the servos (Appendix~\ref{S-app:so101force}).
With randomized layouts and 10 trials per task, \model{} solves every pick, pick-and-place, and ordered-placement trial and 8 of 10 stacking trials (Table~\ref{tab:real}, Figure~\ref{fig:real}a; failures in Appendix~\ref{S-app:so101fail}).
It also follows instructions that require common sense, such as putting the only meat among three ingredients into a bowl (Appendix~\ref{S-app:so101}).

\paragraph{XR1.}
On an XR1 arm with head RGB-D and wrist cameras, \model{} completes tasks with new objects, layouts and indirect object references (Figure~\ref{fig:real}b; Appendix~\ref{S-app:realtrace}).

\subsection{Speed and reaction}
\label{sec:speed}

\begin{figure}[t]
\centering
\includegraphics[width=0.62\linewidth]{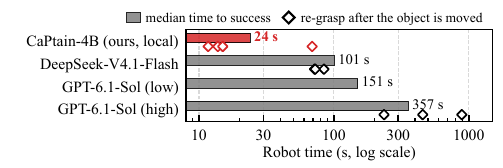}
\caption{\textbf{Speed of \model{} versus frontier models in the Codex harness} on the simulated SO-101 twin (robot time, log scale). Bars: median time to success; diamonds: re-grasp time after the target is moved, one per layout (not run for GPT-6.1-Sol low).}
\label{fig:speed}
\end{figure}

\model{} runs locally on two consumer GPUs at 1.2\,s of robot time per request; quantization and speculative decoding cut model time per request from 5.1\,s to 0.76\,s (Appendix~\ref{S-app:speed}).
On the same tasks and layouts of the simulated SO-101 twin (Figure~\ref{fig:speed}), \model{} completes tasks in a median of 24\,s, against 101\,s for DeepSeek-V4.1-Flash and 151\,s and 357\,s for GPT-6.1-Sol at low and high reasoning effort, 4--15$\times$ faster.
When the target is moved during the first grasp, \model{} re-grasps it within 12--15\,s on three of four layouts (69\,s on the fourth), against 72--85\,s for DeepSeek-V4.1-Flash and 236--890\,s for GPT-6.1-Sol.
Its median time to success ranges from 16.8\,s (pick-and-lift) to 36.6\,s (placing into a container).

%% file: sections/transfer.tex
\subsection{Zero-shot transfer}
\label{sec:transfer}

We evaluate \model{}, trained only on LIBERO, on RoboTwin 2.0~\citep{robotwin2}, SimplerEnv~\citep{simplerenv2024}, CaP-Bench~\citep{fu2026capx} and RoboDojo~\citep{robodojo2026} without further training (Table~\ref{tab:transfer}; adaptation, protocols and details in Appendices~\ref{S-app:adapt}, \ref{S-app:results} and~\ref{S-app:transferdetail}).

\model{} reaches 83.3\% on SimplerEnv and 75.0 on CaP-Bench, outperforming CaP-X with Gemini-3-Pro and the 7B coder trained with CaP-RL on CaP-Bench itself.
The RL gain carries over to new embodiments without target-domain training: RL adds 20 points over \model{}-SFT on CaP-Bench and 4.8 on a separate set of 250 randomized RoboTwin scenes (21.6\% vs.\ 16.8\%), where it also cuts false finishes from 45 to 16.
Without CaP-Gym's step cutoff, \model{} completes 39 of 40 CaP-Bench trials.
Under the official RoboTwin Hard protocol (5{,}000 episodes), \model{} reaches 22.6\% without RoboTwin demonstrations, above X-VLA, $\pi_0$ and RDT trained on them and below $\pi_{0.5}$ (46.0\%).
On RoboDojo's reach-limited ARX X5 arms it scores 1.7 (Appendix~\ref{S-app:robodojo}); most failures come from simultaneous two-arm motion and the limited reach of unfamiliar arms (Appendix~\ref{S-app:failtax}; episodes in Appendix~\ref{S-app:transferep}).

%% file: sections/rest.tex
\section{Conclusion}
\label{sec:conclusion}

We presented \method{}, which post-trains a small open VLM into a robot-use agent over low-level primitives with a \critic{} used only in training.
\model{} improves over SFT on LIBERO and three perturbation suites, transfers to new benchmarks and real robots, and reacts to a changing scene within seconds, suggesting that small, fast, post-trained models are a practical path for robot-use agents.

%% file: sections/appendix.tex
\section{\harness{}}
\label{app:grp-harness}

\subsection{Implementation details}
\label{app:impl}

\subsubsection{Policy and harness}
\begin{figure}[h]
\centering
\includegraphics[width=\linewidth]{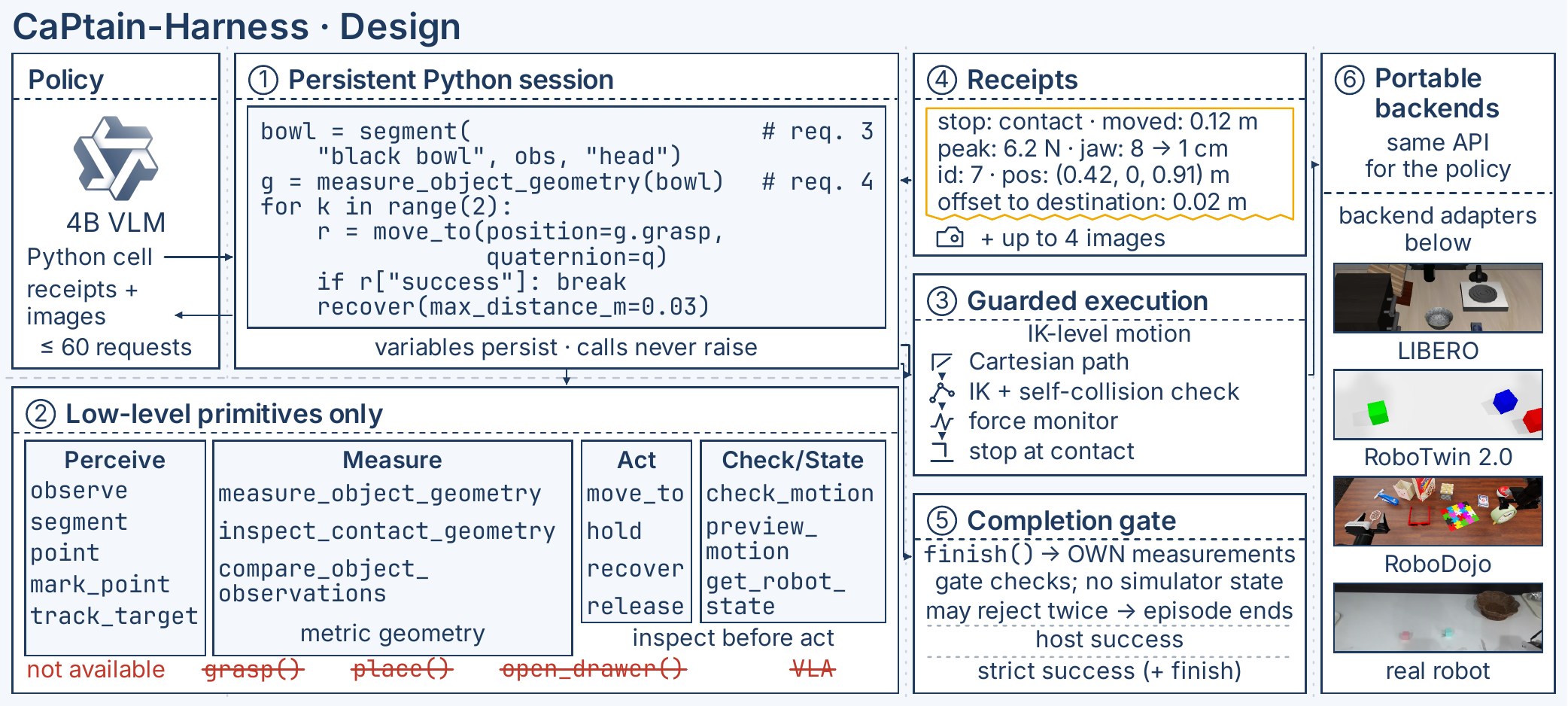}
\caption{\textbf{Design of \harness{}.}
The policy writes Python into a persistent session \ding{172} that exposes only low-level primitives \ding{173}; motions are executed with IK and guarded contact \ding{174}, every call returns a receipt \ding{175}, a completion gate checks the agent's own measurements \ding{176}, and environment-specific backends sit below the API \ding{177}.
Receipt values are illustrative.}
\label{fig:harness}
\end{figure}

\paragraph{Policy.}
The policy $\pi_\theta$ is Qwen3.5-4B~\citep{qwen35blog}: 32 layers with hybrid Gated-DeltaNet and full attention, plus a 24-layer vision tower.
It is served with vLLM~\citep{kwon2023vllm} and sampled at temperature 1 with top-$p$ 1, in training and in evaluation.
Each \emph{request} is one chat completion of at most 4{,}096 tokens with thinking enabled.
The request uses the same chat template (system prompt, developer messages with \texttt{TASK.md}, tool schema) as the SFT trajectories, so training data and policy requests have identical format.
The completion may contain calls to a single tool, \texttt{python}.
The context window is 65{,}536 tokens, and older turns are compacted beyond 60k tokens.

\paragraph{The \texttt{python} tool and receipts.}
The \texttt{python} tool executes code inside a persistent, sandboxed Python 3.11 session (\texttt{numpy}, \texttt{scipy}, \texttt{Pillow}) with a pre-imported \texttt{robot} client module, allowing variables and helper functions to persist across requests.
Robot calls never raise exceptions; instead, each returns a structured dictionary \texttt{\{success, executed, failure\_category, message, value, diagnostics, budget\}} alongside a concise ($\le$300-character) text \emph{receipt} appended to the cell output.
Motion receipts report the stop reason, executed distance, peak contact force, and jaw opening before and after execution, whereas perception receipts report the detection identifier, metric position, and object-to-destination offset.
The policy may attach up to four camera images per cell, with the conversation history retaining the twelve most recent images.

\paragraph{Robot API.}
The API has 22 capabilities plus \texttt{finish()} (full list in Appendix~\ref{app:api}).
They fall into five groups:
\begin{itemize}
\item \textbf{Perception:} \texttt{observe} (head and wrist RGB-D at $512{\times}512$), \texttt{segment} (SAM~3~\citep{sam3}), \texttt{point} (Molmo2-8B~\citep{clark2026molmo2}), \texttt{mark\_point} (back-projects a pixel the model chose itself), \texttt{select\_detection} and \texttt{track\_target}.
\item \textbf{Measurement:} \texttt{measure\_object\_geometry}, \texttt{measure\_image\_points}, \texttt{inspect\_contact\_geometry}, \texttt{inspect\_release\_geometry} and \texttt{compare\_object\_observations}.
\item \textbf{Motion:} \texttt{move\_to}, \texttt{hold}, \texttt{recover} and \texttt{release}. \texttt{move\_to} takes either an absolute end-effector pose or a single jaw command.
\item \textbf{Planning:} the read-only checks \texttt{check\_motion} and \texttt{preview\_motion}.
\item \textbf{State:} \texttt{get\_robot\_state}, \texttt{gripper\_state}, \texttt{robot\_geometry}, \texttt{read\_motion\_trajectory} and \texttt{transform\_pose}.
\end{itemize}
There are no grasp or place macros.
The policy must choose poses, approach directions and jaw commands itself.
Motions are straight Cartesian sweeps executed by an operational-space controller.
Damped least-squares IK and self-collision checks run before execution, and per-substep force monitoring applies a guarded stop at first contact or tolerates load in \texttt{manipulation} mode.
\texttt{TASK.md} tells the policy the instruction, the API, the coordinate frames and the standards by which the final scene is judged.
A host-side completion gate checks only the agent's \emph{own} recorded measurements, never simulator state.
It can reject a \texttt{finish()} twice; a third call is accepted.

\paragraph{Budgets and success.}
An episode may use at most 60 requests, 1{,}080\,s of wall time and 150 physical actions (120 during RL).
\emph{Host success} is the benchmark's goal predicate evaluated on the final state.
\emph{Strict success} additionally requires that the agent ended the episode by calling \texttt{finish()}.
A \emph{false finish} is a \texttt{finish()} on a failed state.
Strict success is the deployable metric: a robot that never declares completion cannot be used.

\subsubsection{RL}
\paragraph{Privileged state.}
Before each request $t$, the environment server records a hidden snapshot that the policy cannot access.
From this mark we compute 36 features $x(s_t)$:
\begin{itemize}
\item end-effector position and jaw state;
\item goal bookkeeping: number of goals, number and fraction satisfied, active goal, and whether that goal involves a joint (articulated part) or contact;
\item distances: target to goal (raw and normalized), end-effector to target, and target to destination;
\item target lift and displacement, maximum displacement of any object, and the nearest-other-object distance;
\item joint position and change for articulated goals, plus \emph{held} and \emph{ever-held} flags.
\end{itemize}
Counters of the policy's own behavior, such as the number of perception calls or failed calls, are excluded so that elapsed time does not leak into the value estimate (Appendix~\ref{app:criticdesign}).

\paragraph{Critic.}
The critic $V_\phi(s,t)=\sigma\!\big(\tfrac{1}{5}\sum_{k=1}^{5} f_{\phi_k}([x(s);\,t;\,\mathrm{onehot}(\text{task})])\big)$ is an ensemble of five MLPs (three hidden layers of 256 units) whose logits are averaged.
It predicts the probability of host success from the current state.
We include the request index $t$ explicitly but hold it fixed inside each difference (below), so elapsed time does not leak into the credit.
The critic is warm-started on 1{,}000 episodes of an earlier policy (45k request states, held-out AUC 0.907, Brier 0.075).
It is then \emph{co-trained} online as in SAO: after each round of 256 episodes it takes two AdamW steps of binary cross-entropy against host success over all request states of that round.
Each round's advantages use the critic \emph{before} that round's update.

\paragraph{Advantage.}
For episode $i$ with requests $t=1,\dots,T$, we define:
\begin{equation}
\delta_t = c_t + V_\phi(s_{t+1},t) - V_\phi(s_t,t)\quad(t<T),\qquad
\delta_T = c_T + R_i - V_\phi(s_T,T),
\label{eq:td-app}
\end{equation}
and $A^{\mathrm{GAE}}_t=\sum_k \lambda^k\delta_{t+k}$ with $\lambda=0.9$ and no discount~\citep{schulman2015gae}.
Each start state is rolled out $G{=}4$ times.
We add a group-relative outcome term:
\begin{equation}
A_{i,t} = \big(G_i - \overline{G}_{\text{group}}\big) + \big(A^{\mathrm{GAE}}_{i,t} - \overline{A^{\mathrm{GAE}}_{i,\cdot}}\big),\qquad G_i = R_i + \textstyle\sum_t c_t ,
\label{eq:adv-app}
\end{equation}
and divide by a running estimate of the advantage standard deviation, without re-centering.
All tokens of request $t$ share the advantage $A_{i,t}$.
Holding $t$ fixed inside each difference prevents elapsed time from leaking into $\delta_t$, while online co-training keeps the value estimate calibrated to the improving policy.
Physical state changes only through physical actions, so the one-step term $V_\phi(s_{t+1},t)-V_\phi(s_t,t)$ is near zero for requests that only perceive, regardless of how many tokens they contain. Such requests still receive credit through GAE from the physical actions they enable, and through the group outcome term.

\paragraph{Returns and costs.}
The terminal return $R_i$ is:
\begin{itemize}
\item $1$ for host success with \texttt{finish()};
\item $0.25$ for success without \texttt{finish()} (a missed finish);
\item $-0.3$ for a false finish;
\item $-0.1$ when a budget is exhausted;
\item $0$ for any other failure.
\end{itemize}
The immediate costs $c_t\le 0$ penalize:
\begin{itemize}
\item physical actions beyond 1.5$\times$ the median action count of the task's SFT episodes ($-0.008$ each);
\item failed or rejected physical actions ($-0.02$ each);
\item actions and requests after the host first judges the scene solved (capped at $-0.45$ in total).
\end{itemize}
The post-success cost pushes the policy to declare completion; without it, the missed-finish rate rose to about 20\% (Appendix~\ref{app:reward}).
For the eight hardest tasks (four from libero\_10 and four from libero\_goal) we add a staged one-shot subgoal reward.
A failed episode that reaches rung $d$ of an $n$-rung ladder (grasp, lift, place, joint opened, \dots) receives $0.15\,d/(n{+}1)\le 0.1125$.
Rungs are judged from hidden snapshots; place rungs and rungs for moving an object toward its goal must still hold at the end.
The resulting ordering is: success $>$ missed finish $>$ partial progress $>$ 0 $>$ budget exhaustion $>$ false finish.

\paragraph{Policy update.}
Following SAO~\citep{sao2026}, we use single-pass, streaming RL with double-sided token masking (unlike SAO, with groups of $G{=}4$ rollouts per start).
A token is kept only if $0.2<\pi_\theta/\pi_{\text{rollout}}<4.0$; it is then weighted by that ratio.
We also apply a k3 KL penalty to the SFT model ($\beta{=}0.002$) on kept tokens.
At most 8 requests per trajectory are trained, sampled uniformly and inclusion-weighted; the critic still uses all requests.
Gradients are accumulated over $K{=}4$ rounds of 256 episodes (64 start groups $\times$ 4), i.e.\ 1{,}024 episodes per optimizer step, matching the empirical gradient-noise scale (Appendix~\ref{app:noise}).
We use AdamW with a constant learning rate of $4.24\times10^{-6}$ after warm-up, full fine-tuning with FSDP2, and at most two policy versions of staleness.

\paragraph{Training distribution.}
RL samples from a pool of 7{,}360 fresh start states over all 40 LIBERO tasks (spatial, object, goal, long), of which 649 libero\_object starts with a degenerate goal region are never drawn:
\begin{itemize}
\item 5{,}600 wide-layout starts with light, camera and material perturbations;
\item 1{,}760 official initial states.
\end{itemize}
The 240 official states used for evaluation (indices 11--16 of each task) are never drawn.
To save rollout time, the critic also stops rollouts early: an episode is ended when $V_\phi<0.02$ for five consecutive requests, except in a 5\% control slice and on the hard tasks. A cut episode settles as budget-exhausted ($R{=}{-}0.1$), so the critic also shapes returns.
A held-out monitor stream (every tenth episode) receives no gradient.

\paragraph{Infrastructure and compute.}
The trainer occupies 8 H100s.
A second 8-GPU node serves 6 vLLM policy servers, SAM~3 and Molmo, and renders 64--80 concurrent MuJoCo episodes.
Weights are synchronized to vLLM by NCCL in under 1\,s.
A round takes 812--1{,}053\,s on 16 GPUs.
Per request, generation takes about 2.6\,s of the 6.7\,s wall time, and the environment, tools and perception take the rest.
The main checkpoint, \model{}, has seen 300 rounds, i.e.\ 75 optimizer updates on 76{,}800 episodes, for 1{,}290 H100-hours of wall time.

\input{sections/appendix_method_extra}

\subsection{Robot API}
\label{app:api}
Table~\ref{tab:api} lists the full \harness{} API used in all experiments: the 22 capabilities and \texttt{finish()}, each with a one-line description. Physical actions count against the action budget; all other calls are free.

\begin{table}[h]
\centering\small
\caption{The 22 capabilities plus \texttt{finish()}. Physical actions (counted against the action budget) are marked $\bullet$.}
\label{tab:api}
\begin{tabular}{@{}lp{0.54\linewidth}@{}}
\toprule
Capability & Semantics \\
\midrule
\texttt{observe} & head + wrist RGB-D ($512^2$), calibration, robot state \\
\texttt{segment} & SAM~3 text segmentation (Molmo point fallback) $\to$ depth-backed detections \\
\texttt{select\_detection} & bind a chosen detection to identity evidence (read-only) \\
\texttt{point} & Molmo pointing $\to$ deprojected, point-segmented evidence \\
\texttt{mark\_point} & validate and back-project a model-chosen pixel $\to$ detection id \\
\texttt{track\_target} & optical-flow tracking of a detection into a newer observation \\
\texttt{read\_motion\_trajectory} & page through the measured pose/joint/force trace of a motion \\
\texttt{get\_robot\_state}, \texttt{gripper\_state} & measured EEF pose, joints, aperture, wrist force \\
\texttt{move\_to}$\bullet$ & absolute EEF pose \emph{or} one jaw command; guarded / manipulation contact mode \\
\texttt{hold}$\bullet$, \texttt{recover}$\bullet$, \texttt{release}$\bullet$ & keep pose; bounded reversal along the last trace; bounded opening under load \\
\texttt{preview\_motion}, \texttt{check\_motion} & read-only timing; IK, self-collision and surface check \\
\texttt{measure\_image\_points}, \texttt{transform\_pose} & metric points at pixels; pose arithmetic \\
\texttt{robot\_geometry} & EEF/pad reference, pad envelope, gripper limits \\
\texttt{inspect\_contact\_geometry} & project the finger-pad closing volume onto the target's RGB-D \\
\texttt{inspect\_release\_geometry} & signed footprint clearance at a proposed release \\
\texttt{compare\_object\_observations} & target motion in robot frame across time \\
\texttt{measure\_object\_geometry} & symmetry-completed center, extents, yaw, object-to-destination offset; part motion \\
\texttt{finish(note)} & end the episode; the host scores the final state (no verdict returned) \\
\bottomrule
\end{tabular}
\end{table}

\subsection{Harness adaptation per transfer benchmark}
\label{app:adapt}
Table~\ref{tab:adapt} lists the backend changes per benchmark and marks each as generic, i.e.\ applicable to any benchmark, or specific to one benchmark or embodiment.
\begin{table}[h]
\centering\small
\caption{Backend changes per transfer benchmark (\emph{Scope}: Generic applies to any benchmark; Benchmark-specific applies to a single benchmark or embodiment).}
\label{tab:adapt}
\begin{tabularx}{\linewidth}{@{}lXc@{}}
\toprule
Benchmark & Change & Scope \\
\midrule
RoboTwin~2.0 & Optional \texttt{arm} argument; a call without it goes to the arm nearer the target & Generic \\
 & Contact detection recalibrated for the Franka gripper & Generic \\
 & Tilted re-approach when IK fails; gripper closes on contact; grasp check reported in the receipt & Generic \\
 & Automatic hand-over for tasks that pass an object between arms & Specific \\
RoboDojo & Success check that does not end the episode (the official check is final) & Generic \\
 & Simulated depth and camera calibration as perception inputs & Specific \\
 & Wrist force estimated from the last arm joint (the X5 has no wrist sensor) & Specific \\
 & Speed limits and IK initialization for the X5 arm & Specific \\
 & Return to the home pose before the official check; automatic grasp retry; a finish instruction in \texttt{TASK.md} & Specific \\
SimplerEnv & Arm joint targets through the robot's own drives instead of the delta end-effector controller; rendered depth and camera parameters, and an added wrist camera, as perception inputs & Specific \\
CaP-Bench & CaP-Gym tasks rebuilt in robosuite~1.4; agent camera at CaP-Gym's pose; wrist flip disabled & Specific \\
\bottomrule
\end{tabularx}
\end{table}

\subsection{Example episode}
\label{app:example}
Excerpt from a strict-success \model{} episode: \emph{``pick up the black bowl between the plate and the ramekin and place it on the plate''} (libero\_spatial task 0, official state 14; 14 requests in total).
Code is verbatim, with long lines and printing trimmed (\texttt{[...]}).
\texttt{go} is the policy's own helper (\S\ref{M-sec:sft}).
Receipts are shortened.
{\scriptsize
\begin{verbatim}
# request 6 (after grasp): lift, re-observe, re-identify bowl and plate by clicking pixels, measure
lift=go(np.array(obs2["value"]["robot_state"]["end_effector_position_m"])+[0,0,.12], q,
        speed=0.035, mode="manipulation")
obs3=robot.observe(); image(obs3["value"]["images"])
b2=robot.mark_point(obs3["value"]["observation_handle"],"agentview",.75,.421,label="held black bowl")
p2=robot.mark_point(obs3["value"]["observation_handle"],"agentview",.78,.725,label="plate")
g2=robot.measure_object_geometry(b2["value"]["detection_id"],p2["value"]["detection_id"])
>>> go: True none bounded Cartesian sweep reached its endpoint  [...]
# request 7: carry the bowl by the measured object-to-destination offset
transfer=np.array(obs3["value"]["robot_state"]["end_effector_position_m"])
transfer[:2]-=np.array(g2["value"]["object_to_destination_xy_offset_m"])
trans=go(transfer, q, mode="manipulation")
>>> move_to: ok; stopped_reason=target_reached executed_distance_m=0.1681 max_observed_force_n=5.198
# requests 8-10: head view is occluded by the held bowl -> re-measure from the wrist camera
p3w=robot.mark_point(obs4["value"]["observation_handle"],"robot0_eye_in_hand",.53,.58,label="plate")
b3w=robot.mark_point(obs4["value"]["observation_handle"],"robot0_eye_in_hand",.80,.62,label="black bowl")
g3w=robot.measure_object_geometry(b3w["value"]["detection_id"],p3w["value"]["detection_id"])
# request 11: open, retreat, settle, leave a note for itself
opened=robot.move_to(gripper="open")
retreat=go(release_pose+np.array([0,0,.14]), q, speed=0.035); settled=robot.hold(duration_s=1.0)
>>> move_to: ok; gripper_open_fraction=0.04354->0.9904 ... hold: ok
# request 13: fresh final measurement
pf2=robot.mark_point(obs6["value"]["observation_handle"],"agentview",.70,.74,label="plate",
                     granularity="object")
finalg2=robot.measure_object_geometry(bf["value"]["detection_id"],pf2["value"]["detection_id"])
>>> {'object_to_destination_xy_offset_m': [0.0012, 0.0170], 'xy_distance_m': 0.0171,
     'resting_height_consistent': True}
"The bowl is resting on the plate. Fresh measurements show a center offset of about 1.7 cm,
 within the required 2 cm tolerance."   -> robot.finish(...)   [host: success, agent_finished]
\end{verbatim}
}
The episode shows three behaviors of the RL policy.
It grounds objects by clicking pixels and back-projecting them rather than calling text segmentation.
When the held object occludes the head camera, it switches to the wrist camera.
It verifies the final relation with a fresh measurement against the \texttt{TASK.md} tolerance before calling \texttt{finish()}.

\section{Training}
\label{app:grp-training}

\subsection{SFT details}
\label{app:sft}
\begin{table}[h]
\centering\small
\caption{\model{}-SFT data mix (25{,}961 rows, 861M tokens). Generating agents: GPT-6 Astra (654M tokens), GPT-6.1 Sol (63M), GPT-6 Sol (14M); RL policy (52M).}
\begin{tabular}{@{}lrrr@{}}
\toprule
Source & Rows & Tokens & Share \\
\midrule
Broad-coverage trajectories (code-normalized, filtered) & 7{,}903 & 348.9M & 40.5\% \\
Hard-task targeted trajectories (Round 2) & 5{,}090 & 238.4M & 27.7\% \\
Hard-task targeted trajectories (Round 1) & 2{,}138 & 99.3M & 11.5\% \\
Hard-task trajectories (60-request cap) & 925 & 43.4M & 5.0\% \\
Self-generated RL rollouts (successful episodes) & 1{,}386 (267 episodes) & 52.3M & 6.1\% \\
Visual QA (loss weight 0.5) & 8{,}519 & 78.8M & 9.2\% \\
\bottomrule
\end{tabular}
\end{table}

\paragraph{Data generation.}
The synthetic episodes were generated by frontier coding agents (GPT-6 Astra and GPT-6.x Sol, see the table caption) acting in the same harness, with the same API, \texttt{TASK.md} and chat template as the 4B policy.
The raw episodes contained 226 distinct variants of the same ``preview, then move'' helper.
We rewrote all trajectories onto a single helper library of four functions (\texttt{val}, \texttt{need}, \texttt{downq}, \texttt{go}) that the model learns to write itself.
Equivalence of each rewrite was checked call site by call site against a mock robot, and replaying 6 rewritten episodes changed robot commands by less than 1\,cm.
Turns with repeated execution failures or recovery loops (about 3.6\% of requests) are masked from the loss, as are requests after a gate-rejected finish on one long-horizon task.

\paragraph{Training.}
Only assistant turns are supervised.
A text-hint mask replaces earlier \texttt{mark\_point} pixel coordinates in the history with \texttt{<masked>} with probability 0.2.
This forces the model to look at the image instead of copying coordinates.
Training uses full fine-tuning of the language model with the vision tower frozen.
The batch is 16 packed rows of at most 66k tokens.
We use DeepSpeed ZeRO-3 in bf16, gradient clipping at 1.0, and 2{,}435 steps (1.5 epochs).
The final model scores 42/80 on our 80-episode validation set.

\subsection{Diversity of tasks and layouts during post-training}
\label{app:diversity}
Diversity of tasks and scene layouts in the post-training data is essential for keeping the visual grounding of the base model.
In the official LIBERO initial states, each object occupies nearly the same position across episodes (per-object position standard deviation about 1.4\,cm).
Models fine-tuned on SFT data that start mostly from these states learned to reproduce object positions from the task identity rather than reading them from the image: replacing the input image left the likelihood of the predicted pixel coordinates unchanged.
The policy had effectively memorized the layouts and lost the vision-language model's ability to ground objects in new scenes.
We therefore collect SFT data on wide-layout starts, whose per-object position standard deviation is about 15\,cm, mix all task suites, and randomize lighting, camera pose and materials during RL.
With this diversity, the \texttt{mark\_point} coordinates depend on the current image (Appendix~\ref{app:interp}).

\subsection{Reward, costs and early stopping}
\label{app:reward}
The post-success cost (every physical action at $-0.008$ plus $-0.005$ per request after the host first judges success, capped at $-0.45$) reduces missed finishes, which rose to about 20\% per round without it.
The action costs counter the growth of \texttt{move\_to} calls per request under RL: from 0.69 to 1.05 in successful episodes and from 0.78 to 1.39 in failed ones.
The request cap did not constrain the number of actions packed into one request.
The staged subgoal reward is a one-shot, monotone ladder judged from hidden snapshots.
\begin{itemize}
\item It applies to 8 hard tasks (libero\_10 3/6/8/9; goal 0/3/5/7), from round 9 onward.
\item A grasp rung requires the object lifted by at least 5\,cm with the end-effector within 13\,cm.
\item A joint rung requires the articulated part to cover at least 20\% of its initial distance to the goal position.
\end{itemize}
The critic also stops rollouts early, for speed.
\begin{itemize}
\item An episode is cut when $V<0.02$ for 5 consecutive requests after request 5. A deterministic 5\% control slice is never cut.
\item An episode is stopped and settled as a missed finish once all goals have held for 15 consecutive requests without \texttt{finish()}.
\end{itemize}

\subsection{RL infrastructure}
\label{app:infra}
\paragraph{Layout.}
Rollout and training run on separate 8-GPU nodes.
The rollout node serves six vLLM policy replicas, SAM~3 and Molmo, and renders 64--80 concurrent MuJoCo episodes; the trainer uses FSDP2 on the other node.
Rollouts are streamed: the environment server keeps generating while the trainer updates, and a request may come from a policy that is at most two versions old.
New weights reach the rollout servers by NCCL in under one second, and checkpoints are written asynchronously, which reduces the blocking time of a checkpoint from about 500\,s to 5--9\,s.

\paragraph{Speed-ups.}
Table~\ref{tab:infra} lists the optimizations we kept.
Together they shortened a training round by 31\% (1{,}812\,s$\to$1{,}246\,s on the same benchmark round), after which training became trainer-bound.
\begin{table}[h]
\centering\small
\caption{\textbf{RL infrastructure optimizations} that passed the numerical gates below.}
\label{tab:infra}
\begin{tabularx}{\linewidth}{@{}>{\raggedright\arraybackslash}X>{\raggedright\arraybackslash}X@{}}
\toprule
Optimization & Effect \\
\midrule
FlashInfer attention and a chunked-prefill budget of 8{,}192 tokens in vLLM & +17.6\% policy throughput (30.4$\to$35.7 requests per minute per GPU); log-probabilities within the kernel noise floor \\
Sensor artifacts in shared memory & \texttt{observe} latency 2.2\,s$\to$0.16\,s \\
Harness fast loop and lazy rendering & $-27\%$ episode wall time, bit-identical physics and responses \\
Critic-based early stopping (5\% control slice never stopped) & 24--26\% of episodes cut early; no successful episode in the control slice would have been cut \\
KL reference evaluated every 8 steps & unbiased KL estimate at $1/8$ of the reference cost \\
Pack-level rank assignment & balanced trainer ranks; identical log-probabilities, gradient cosine 0.9999 \\
\bottomrule
\end{tabularx}
\end{table}

\paragraph{Numerical gates.}
Every speed-up had to keep the trained objective unchanged: per-token log-probabilities and batch gradients of the new path had to stay within the difference between two runs of the reference path.
The trainer uses FlashAttention-2 / SDPA.

\subsection{Compute accounting}
\label{app:compute}
\begin{itemize}
\item \textbf{RL rounds.} A round takes 812\,s before the staged reward and 991--1{,}053\,s after it on 16 H100s.
\item \textbf{Rollout supply.} 873--1{,}135 episodes per hour.
\item \textbf{Per request.} About 6.7\,s, of which 2.6\,s is in vLLM (0.3\,s prefill, 2.1\,s decode).
\item \textbf{Trainer.} About 0.53\,s per packed sequence and 1{,}269--1{,}507 sequences per round. The time splits into 24\% forward, 72\% backward and 3\% reference forward.
\item \textbf{Weight sync and checkpoints.} Weights are synchronized to all vLLM servers over NCCL in 0.8\,s, and exporting a checkpoint takes 25\,s.
\item \textbf{Totals.} 1{,}290 H100-hours for RL to \model{} (300 rounds) and 120 H100-hours for SFT (wall time $\times$ GPUs). The critic warm start uses 16 GPU-hours of rollouts.
\end{itemize}

\subsection{Lessons from RL on a multi-turn agent}
\label{app:lessons}
\begin{itemize}
\item \textbf{Measure the gradient noise before choosing the batch.}
With two start states per optimizer step, the weight displacement from the SFT model grew as the square root of the number of steps and consecutive updates were nearly orthogonal: the policy performed a random walk.
The gradient-noise scale of outcome-only advantages is in the thousands of episodes (Appendix~\ref{app:noise}), so we accumulate 1{,}024 episodes per optimizer step.
\item \textbf{Start states must cover the whole task.}
A curriculum that started episodes at intermediate states of successful demonstrations placed every start after the grasp, because a grasp does not change the distance of the object to its goal; the policy then learned only to place held objects.
We train from fresh initial states only.
\item \textbf{Layout diversity keeps the model looking.}
Without diverse layouts, the model memorizes object positions and stops grounding its calls in the image (Appendix~\ref{app:diversity}).
\item \textbf{Keep the policy's behavior out of the critic.}
Counters of perception or failed calls act as proxies for elapsed time and move credit from grasps to perception (Appendix~\ref{app:criticdesign}).
\item \textbf{Termination needs its own signal.}
A critic trained on task success values every post-success state near one, so the decision to stop receives no credit from the critic; post-success costs and a missed-finish return are needed to make the agent declare completion (Appendix~\ref{sec:collapse}).
\end{itemize}

\subsection{Gradient-noise analysis}
\label{app:noise}
\begin{table}[h]
\centering\small
\caption{Gradient-noise scale $B_{\text{noise}}$ (episodes) of advantage estimators on 109 groups $\times$ 16 SFT rollouts (offline; last four layers sketched). Lower is better.}
\begin{tabular}{@{}lc@{}}
\toprule
Estimator & $B_{\text{noise}}$ \\
\midrule
GRPO (0/1 outcome) & 9.0k \\
\critic{}, $\lambda{=}0$ & 12.6k \\
\critic{}, $\lambda{=}0.9$ & 6.7k \\
\critic{}, $\lambda{=}1$ & 6.5k \\
\bottomrule
\end{tabular}
\end{table}
For the 4B model, the simple noise scale at an early checkpoint was about 9{,}500 episodes (95\% lower bound 3{,}300).
This motivated accumulating $K{=}4$ rounds (1{,}024 episodes) per update.
Without accumulation, consecutive updates had near-zero cosine similarity.

\section{RL analysis}
\label{app:grp-rl}

\subsection{Training curves of the RL ablations}
\label{app:ablcurves}
\begin{figure}[h]
\centering
\includegraphics[width=\linewidth]{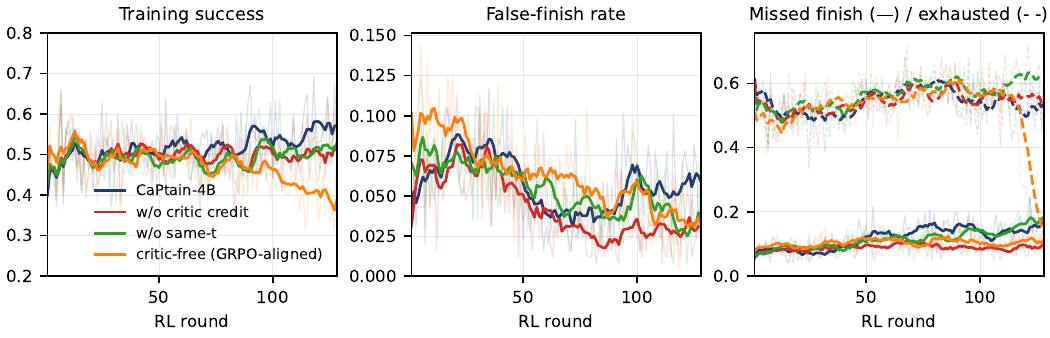}
\caption{\textbf{Training curves of the RL ablations} (Table~\ref{M-tab:ablation}), rounds 1--128. Per-round host success of the 256 training rollouts, false-finish rate, and missed-finish (solid) and budget-exhausted (dashed) rates; thick lines are 8-round moving averages.}
\label{fig:ablcurves}
\end{figure}

\begin{table}[h]
\centering\small
\caption{\textbf{Training success of the RL ablations} (\%, per-round host success of the 256 training rollouts, averaged over round windows). $\Delta$ is the matched-round difference to \model{}. Brackets: 95\% moving-block bootstrap intervals over rounds (block 8, 10{,}000 resamples); they reflect round-to-round noise of one seed per run, not seed-to-seed variance.}
\label{tab:ablcurves}
\begin{tabular}{@{}lcccc@{}}
\toprule
Run & Rounds 1--50 & Rounds 51--100 & Rounds 101--128 & $\Delta$ rounds 101--128 \\
\midrule
\model{} & 49.9 [48.9, 51.3] & 52.3 [50.6, 54.1] & 55.8 [53.1, 57.6] & -- \\
w/o critic credit & 50.1 [49.1, 52.1] & 50.0 [48.5, 51.5] & 50.3 [48.8, 51.3] & $-5.4$ [$-7.2$, $-3.8$] \\
w/o same-$t$ & 49.2 [48.3, 51.0] & 49.0 [47.0, 50.9] & 51.5 [49.8, 52.1] & $-4.3$ [$-6.5$, $-2.2$] \\
critic-free (GRPO) & 50.0 [48.4, 51.9] & 47.7 [46.3, 49.0] & 40.6 [38.8, 42.1] & $-15.1$ [$-18.1$, $-11.3$] \\
\bottomrule
\end{tabular}
\end{table}

All runs are indistinguishable over the first 50 rounds. The gaps open later: \model{} still improves from rounds 51--100 to 101--128 ($+3.4$ points, 95\% CI $[+0.3,+6.0]$), the two critic ablations plateau, and the critic-free run declines ($-7.0$, $[-9.3,-5.0]$); its 8-round moving average peaks at round 13 (55.9\%) and ends at 38.5\%.

\subsection{Collapse of the policy value head}
\label{app:vbcollapse}

\begin{table}[h]
\centering\small
\caption{\textbf{Training rollouts of the policy-value-head ablation} (256 per round). \emph{Stopped}: the policy ends the episode without \texttt{finish()}; \emph{exhausted}: the request budget runs out; $\bar V$: mean predicted success of the value head; $\bar A$: mean advantage before normalization.}
\label{tab:vbcollapse}
\setlength{\tabcolsep}{4pt}
\begin{tabular}{@{}rcccccccc@{}}
\toprule
Round & Success & \texttt{finish()} & False finish & Exhausted & Stopped & Requests/ep. & $\bar V$ & $\bar A$ \\
\midrule
20 & 0.570 & 154 & 22 & 102 & 0 & 17.6 & 0.79 & $+0.05$ \\
65 & 0.469 & 107 & 11 & 149 & 0 & 18.1 & 0.76 & $+0.04$ \\
80 & 0.512 & 113 & 11 & 111 & 32 & 20.0 & 0.82 & $+0.03$ \\
92 & 0.375 & 67 & 1 & 2 & 187 & 10.6 & 0.64 & $-0.01$ \\
100 & 0.113 & 13 & 0 & 0 & 243 & 5.9 & 0.30 & $-0.17$ \\
128 & 0.000 & 0 & 0 & 0 & 255 & 1.7 & 0.01 & $-0.41$ \\
\bottomrule
\end{tabular}
\end{table}

Table~\ref{tab:vbcollapse} traces the run.
Until round 65 it behaves like the other ablations (training success 0.46--0.58).
From round 70, failing episodes move to the one exit with zero return: budget exhaustion returns $-0.1$ and a false finish $-0.3$, but stopping without \texttt{finish()} returns 0 and also avoids the per-action costs.
Between rounds 70 and 92 exhausted episodes fall from 150 to 2 per round and false finishes to 1, while stopped episodes rise from 0 to 187; success still holds at 0.38--0.51.
After round 92 the exit spreads to winnable episodes: \texttt{finish()} calls fall from 67 to 0, requests per episode from 10.6 to 1.7, and the predicted success and the mean advantage fall to near 0 and $-0.4$.
At round 128 the policy writes one sentence of intent and repeats it until the 4{,}096-token limit; all 240 held-out episodes end this way.
The KL to the SFT policy stays between 0.02 and 0.045 throughout, so this is a slow drift of the objective, not an optimization instability.

The value head predicts 0/1 success, so $A_t=R-V(s_t)$ contains no term for costs and penalties, and among failing episodes the one that stops earliest receives the least negative advantage.
The return ordering alone does not cause the collapse: the two ablations that keep the \critic{} (w/o critic credit, w/o same-$t$) use the same returns and show no such drift (0--1 stopped episodes per round through round 128).
The critic-free GRPO run, which shares the returns but has no \critic{}, drifts the same way later: stopped episodes rise from 2 per round at round 80 to 99 at round 120.
Both runs without the \critic{} learn the cheapest failure; neither ablation that keeps it does.

\subsection{Why the critic sees only physical state}
\label{app:criticdesign}

Credit statistics are computed on 18.8k episodes (596k request states) from 16 four-round windows spread over the RL run of \model{}; for comparison, the 9B values below come from offline rollouts of an earlier 9B policy (15k episodes).
On the 4B run the critic reaches AUC 0.962, within-group AUC 0.874 and calibration error 0.008 (0.970, 0.904 and 0.006 on the 9B rollouts); the warm-start critic used for 4B RL reaches 0.907 held-out AUC.
The median $\Delta V$ of a successful grasp on the 4B run is $+0.077$.

Giving the critic counters of the policy's own actions raises within-group AUC on pre-grasp states from 0.773 to 0.862, but the counters act as proxies for elapsed time and leak negative credit onto perception requests.
A linear probe on the policy's hidden state decodes the request index ($R^2$ 0.69 at layer 12; Appendix~\ref{app:interp}), so a critic that reads hidden states can leak time into credit in the same way.
Privileged physical state with a fixed request index avoids this leak, and lowers the gradient-noise scale on 9B SFT rollouts from 9.0k episodes under outcome-only GRPO to 6.7k ($\lambda{=}0.9$; Appendix~\ref{app:noise}).

\subsection{Host and strict success late in RL}
\label{sec:collapse}
\begin{figure}[t]
\centering
\includegraphics[width=0.55\linewidth]{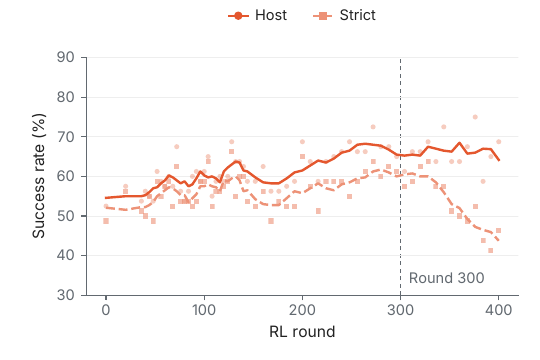}
\caption{\textbf{Finish collapse.} Evaluation of saved checkpoints during RL: 40 official seed-11 episodes plus 40 wide-layout episodes per checkpoint; dots are raw values and lines a 5-checkpoint moving average. After about round 340, host success stays flat while strict success falls.}
\label{fig:collapse}
\end{figure}

Host and strict success diverge late in training (Figure~\ref{fig:collapse}).
After about round 340, host success holds at about 65\% while strict success falls from about 60\% to 44\%.
The lost episodes run out of requests while re-verifying solved scenes, while false finishes stay rare.
Because strict success drops between rounds 300 and 400, we use the round-300 checkpoint as \model{}.

The mechanism is over-verification.
After the task is solved, the policy keeps measuring, re-observing and comparing instead of calling \texttt{finish()}.
The return structure makes this a slow drift rather than a cliff:
\begin{itemize}
\item a missed finish still earns 0.25;
\item the post-success cost is capped;
\item the critic, trained on host success, assigns $V{\approx}1$ to every post-success state.
\end{itemize}
The critic therefore gives the decision to stop no credit; only the terminal return and the small post-success request cost distinguish it.
This suggests giving the decision to finish its own target, e.g.\ success-and-finished.

\input{sections/appendix_interp}

\section{LIBERO results}
\label{app:grp-libero}

\subsection{Additional LIBERO results}
\label{app:results}

\paragraph{Evaluation protocol.}
LIBERO-PRO is reported as the 3-suite average over Spatial, Object and Goal (600 episodes), the 4-suite average adding Long (800 episodes), and the 4-category split (160 episodes).
LIBERO-MAX is evaluated under static (Base) and in-episode (Dynamic) perturbations.
All evaluations run in \harness{} with the RL rollout settings: temperature 1, thinking on, at most 60 requests, 150 physical actions and 1{,}080\,s per episode, and no early stopping.
Comparisons between our models are paired on identical (suite, task, initial state) triples.
Baseline numbers are taken from the respective papers.

\begin{table}[!ht]
\centering\small
\caption{\textbf{LIBERO Held-Out (240 Episodes) Breakdown.} Episode counts out of 240 overall, and per-suite Host/Strict counts out of 60.}
\setlength{\tabcolsep}{3.5pt}
\begin{tabular}{@{}lccccccccc@{}}
\toprule
Model & Host & Strict & Req.\ Cap & False Finish & Missed Finish & Spatial & Object & Goal & Long \\
\midrule
\model{}-SFT & 133 & 125 & 96 & 17 & 8 & 47/46 & 46/45 & 25/21 & 15/13 \\
RL, round 128 & 174 & 159 & 68 & 10 & 15 & 54/54 & 59/57 & 35/31 & 26/17 \\
RL, round 292 & 177 & 156 & 80 & 4 & 21 & 56/56 & 57/51 & 38/35 & 26/14 \\
\rowcolor{oursbg} \model{} (RL, round 300) & 177 & 162 & 73 & 5 & 15 & 58/58 & 55/52 & 41/36 & 23/16 \\
\bottomrule
\end{tabular}
\end{table}
Per-suite paired gains of \model{} over SFT (host): spatial +18.3 ($p{=}10^{-3}$), object +15.0 ($p{=}0.035$), goal +26.7 ($p{=}1.4{\times}10^{-4}$), long +13.3 ($p{=}0.13$).
The libero\_10 task 5 predicate has a 3.5\,mm offset that rejects correct flat placements; every model scores 0/6 on it.
Raising the request cap from 60 to 120 on libero\_10 did not change success (17 vs.\ 18/60 for an earlier checkpoint).
On LIBERO-PRO Pos+Task, a clean subset of 78 episodes per model uses initial state 11, which was never used in training.
It gives the same picture: \model{} reaches 74.6\% (3 suites) and 66.4\% (all 4 suites) host success vs.\ SFT at 57.8\% and 47.4\%.

\begin{table}[!ht]
\centering\small
\caption{\textbf{Exact two-sided McNemar $p$-values} for paired comparisons between \model{} and \model{}-SFT across all LIBERO benchmark suites (corresponding to Table~\ref{tab:main}).}
\label{tab:mcnemar}
\begin{tabular}{@{}lrcc@{}}
\toprule
Benchmark & $n$ & Host $p$-value & Strict $p$-value \\
\midrule
LIBERO & 240 & $4{\times}10^{-8}$ & $7{\times}10^{-6}$ \\
LIBERO-PRO (3 suites) & 600 & $1{\times}10^{-20}$ & $2{\times}10^{-14}$ \\
LIBERO-PRO (All 4 suites) & 800 & $8{\times}10^{-28}$ & $1{\times}10^{-16}$ \\
LIBERO-Plus & 700 & $8{\times}10^{-21}$ & $5{\times}10^{-15}$ \\
LIBERO-PRO (4 categories) & 160 & $4{\times}10^{-7}$ & $3{\times}10^{-3}$ \\
LIBERO-MAX Dynamic & 80 & $2{\times}10^{-3}$ & $0.15$ \\
LIBERO-MAX Base & 80 & $2{\times}10^{-3}$ & $4{\times}10^{-3}$ \\
\bottomrule
\end{tabular}
\end{table}

\paragraph{Baseline protocol notes (Tables~\ref{M-tab:baselines} and~\ref{M-tab:transfer}).}
In Table~\ref{M-tab:baselines} (LIBERO-PRO Pos+Task): OpenRUA~\citep{openrua2026} uses latched success and a wider turn budget (100--500 turns); Harness VLA~\citep{harnessvla2026}, K1~\citep{k1harness2026}, and RATs~\citep{rats2026} use test-time oracle signals, play experience, or memory collected on other seeds of the evaluated tasks; WAA~\citep{waa2026} and SPARK~\citep{spark2026} use privileged simulator geometry or ground-truth depth; ASPIRE~\citep{aspire2026}, SPARK, CaP-Agent0~\citep{fu2026capx}, and $\pi_{0.5}$~\citep{liberopro2025} evaluate 50 trials per task; SPARK and RACaP~\citep{racap2026} run at temperature $T{=}0$; K1 and RACaP evaluate 30 episodes per cell; and AgenticRoboRL~\citep{agenticroborl2026} uses a 100--500 turn budget (its 6-cell average is recomputed from the paper's per-cell numbers).
In Table~\ref{M-tab:transfer} (zero-shot transfer): on RoboTwin~2.0, PyRUA-Lean Coder-4B~\citep{pyrualean2026} evaluates 250 randomized scenes with latched success on native Aloha-AgileX hardware using a RoboTwin-trained VLA tool, whereas \model{} and the VLA/WAM baselines follow the official Hard protocol (100 trials per task across 50 tasks; single-arm VLA numbers are averaged over the 33 single-arm tasks from the official leaderboard~\citep{robotwin2,robotwinboard,openwam2026pretrain}); on RoboDojo~\citep{robodojoboard}, baselines run on official layouts while \model{} is evaluated on 15 layouts per task across all 42 tasks (single-arm GPT-6 Astra is averaged over the 32 single-arm tasks, reaching 31.9 Score / 26.8\% SR); on SimplerEnv, PhysBrain~1.0 and $\pi_{0.5}$ run under the official VLA step limit~\citep{physbrain2026}; and on CaP-Bench, CaP-X (multi-turn) corresponds to the M2 visual-feedback setting while CaP-RL and CaP-X (single-turn) run in the open-loop S2 setting~\citep{fu2026capx}.

\input{sections/appendix_exp_extra}

\subsection{Full LIBERO-family comparison}
\label{app:fullcompare}
Table~\ref{tab:fullcompare} collects published results on standard LIBERO, LIBERO-PRO, LIBERO-Plus and LIBERO-MAX for vision-language-action models (VLAs), world-action models (WAMs) and robot-use agents (RUAs), next to our models.
Most VLAs and WAMs are fine-tuned on LIBERO demonstrations and reach 94--99\% on standard LIBERO, but they retain only 20--87\% of that success under the perturbations of LIBERO-Plus. \model{}, trained without any LIBERO demonstration, retains 90\%.
Baseline numbers are copied from the cited papers; ``--'' marks benchmarks a method has not reported.

\begin{table}[h]
\centering
\caption{\textbf{LIBERO-family comparison} (success \%). \emph{Frontier}: a frontier model is queried while the robot acts. \emph{Demos}: trained on LIBERO demonstrations, directly or through a VLA skill. Bold: best among methods that use neither a frontier model at test time nor LIBERO demonstrations. \emph{Plus / LIBERO}: LIBERO-Plus success as a percentage of standard LIBERO success, i.e.\ how much performance is retained under perturbation. Protocol differences are listed below the table.}
\label{tab:fullcompare}
{\footnotesize
\setlength{\tabcolsep}{3.2pt}
\begin{tabularx}{\linewidth}{@{}>{\raggedright\arraybackslash}X l c c c c c c c c c@{}}
\toprule
\multirow{2}{*}{Method} & \multirow{2}{*}{Family} & Frontier & LIBERO & \multirow{2}{*}{LIBERO} & \multicolumn{2}{c}{LIBERO-PRO} & LIBERO- & \multicolumn{2}{c}{LIBERO-MAX} & Plus / \\
\cmidrule(lr){6-7}\cmidrule(lr){9-10}
 & & at test & demos & & 3 Suites & 4 Suites & Plus & Base & Dyn. & LIBERO \\
\midrule
\rowcolor{oursbg} \textbf{\model{}} (ours) & RUA & \ding{55} & \ding{55} & \textbf{73.8} & \textbf{78.5} & \textbf{68.6} & \textbf{66.1} & \textbf{70.0} & \textbf{61.2} & \textbf{90} \\
\rowcolor{oursbg} \model{}-SFT (ours) & RUA & \ding{55} & \ding{55} & 55.4 & 59.5 & 49.2 & 48.6 & 51.2 & 40.0 & 88 \\
\rowcolor{oursbg} Qwen3.5-4B (base) & RUA & \ding{55} & \ding{55} & 0.0 & 0.3 & 0.2 & 0.0 & 0.0 & 0.0 & -- \\
\midrule
\multicolumn{11}{@{}l}{\emph{Vision-language-action models}} \\
OpenVLA~\citep{kim2024openvla} & VLA & \ding{55} & \ding{51} & 76.5 & 0.0 & 0.0 & 15.6 & -- & -- & 20 \\
OpenVLA-OFT~\citep{kim2025oft} & VLA & \ding{55} & \ding{51} & 97.1 & -- & -- & 69.6 & 64.3 & 43.2 & 72 \\
RIPT-VLA~\citep{riptvla} & VLA & \ding{55} & \ding{51} & 97.5 & -- & -- & 68.4 & -- & -- & 70 \\
$\pi_0$~\citep{black2024pi0} & VLA & \ding{55} & \ding{51} & 94.2 & 0.0 & 0.3$^\dagger$ & 53.6 & -- & -- & 57 \\
$\pi_0$-FAST~\citep{pi0fast} & VLA & \ding{55} & \ding{51} & 85.5 & -- & -- & 61.6 & -- & -- & 72 \\
$\pi_{0.5}$~\citep{pi05} & VLA & \ding{55} & \ding{51} & 96.9 & 12.8$^d$ & 10.8 & 84.4$^\dagger$ & 79.7 & 65.7 & 87 \\
UniVLA~\citep{univla} & VLA & \ding{55} & \ding{51} & 95.2 & -- & 2.5 & 42.9 & -- & -- & 45 \\
GR00T N1.6 / N1.7$^j$~\citep{gr00tn1} & VLA & \ding{55} & \ding{51} & 97.0$^\dagger$ & -- & -- & -- & 69.3 & 50.0 & -- \\
MolmoAct2~\citep{molmoact2} & VLA & \ding{55} & \ding{51} & -- & 18.6$^\dagger$ & -- & -- & 80.3 & 66.9 & -- \\
\midrule
\multicolumn{11}{@{}l}{\emph{World-action models}} \\
Cosmos-Policy~\citep{cosmospolicy} & WAM & \ding{55} & \ding{51} & 98.5 & -- & -- & 82.2$^\dagger$ & 77.4 & 59.3 & 83 \\
Fast-WAM~\citep{fastwam} & WAM & \ding{55} & \ding{51} & 97.6$^\dagger$ & -- & -- & 51.5$^\dagger$ & 42.0 & 24.0 & 53 \\
GE-Act + Robust-WAM~\citep{robustwam2026} & WAM & \ding{55} & \ding{51} & 97.3 & -- & -- & 80.9 & -- & -- & 83 \\
OpenWAM-$\alpha$~\citep{openwam2026pretrain} & WAM & \ding{55} & \ding{51} & 99.3 & -- & -- & 69.2 & -- & -- & 70 \\
\midrule
\multicolumn{11}{@{}l}{\emph{Robot-use agents}} \\
OpenRUA~\citep{openrua2026} & RUA & \ding{51} & \ding{55} & -- & 91.8$^{a,g}$ & 87.0 & -- & -- & -- & -- \\
Harness VLA (Opus 4.7)~\citep{harnessvla2026} & RUA & \ding{51} & \ding{51} & 96.0 & 87.7$^b$ & 82.4 & -- & -- & -- & -- \\
AgenticRoboRL~\citep{agenticroborl2026} & RUA & \ding{55} & \ding{51} & -- & 81.3$^{g,h,i}$ & 71.0 & 72.2 & -- & -- & -- \\
K1~\citep{k1harness2026} & RUA & \ding{51} & \ding{55} & -- & 77.2$^{b,f}$ & -- & -- & -- & -- & -- \\
WAA~\citep{waa2026} & RUA & \ding{51} & \ding{55} & -- & 75.6$^c$ & -- & -- & -- & -- & -- \\
ASPIRE~\citep{aspire2026} & RUA & \ding{51} & \ding{55} & -- & 71.7$^{b,d}$ & -- & -- & -- & -- & -- \\
GPT-6 Astra in \harness{} & RUA & \ding{51} & \ding{55} & -- & 68.5 & -- & -- & -- & -- & -- \\
SPARK~\citep{spark2026} & RUA & \ding{55} & \ding{55} & -- & 66.8$^{c,d,e}$ & -- & -- & -- & -- & -- \\
RACaP~\citep{racap2026} & RUA & \ding{51} & \ding{55} & -- & 45.0$^{e,f}$ & -- & -- & -- & -- & -- \\
Guava~\citep{guava2026} & RUA & \ding{55} & \ding{55} & -- & $\approx$41 & -- & -- & -- & -- & -- \\
CaP-Agent0~\citep{fu2026capx} & RUA & \ding{51} & \ding{55} & -- & 18.2$^d$ & -- & -- & -- & -- & -- \\
\bottomrule
\end{tabularx}}
\par\smallskip
\parbox{\linewidth}{\scriptsize
\emph{Protocols.} Ours: host success; LIBERO on the held-out official initial states 11--16 (240 episodes); LIBERO-PRO Pos+Task with initial states 1--10, 100 episodes per cell; LIBERO-Plus on a stratified 700-case sample; LIBERO-MAX on the 80 Lite cases with independent Base and Dynamic rollouts.
Baselines: LIBERO with 500 episodes per suite (Harness VLA: 100); LIBERO-Plus on the full 10{,}030 instances; LIBERO-MAX on 8{,}000 pairs, with Dynamic replaying the Base prefix.
$^a$latched success;
$^b$test-time success oracle, or memory or programs built on other seeds of the evaluated tasks;
$^c$privileged simulator geometry or ground-truth depth;
$^d$50 trials per task;
$^e$$T{=}0$;
$^f$30 episodes per cell;
$^g$budget of 100--500 turns;
$^h$six-cell average recomputed from the per-cell numbers;
$^i$planner trained on the LIBERO-PRO tasks with other initial states;
$^j$LIBERO for N1.6, LIBERO-MAX for N1.7;
$^\dagger$evaluated by a later paper: \citet{harnessvla2026} ($\pi_0$ PRO 8c), \citet{openwam2026pretrain} (LIBERO-Plus of $\pi_{0.5}$, Cosmos-Policy and Fast-WAM; LIBERO of GR00T N1.6 and Fast-WAM), \citet{spark2026} (MolmoAct2).}
\end{table}

\section{Zero-shot transfer}
\label{app:grp-transfer}

\subsection{Per-benchmark transfer results}
\label{app:transferdetail}

\paragraph{RoboTwin~2.0.}
Under the Hard protocol (5{,}000 episodes), \model{} succeeds in 22.6\% of episodes on all 50 tasks and 30.0\% on the single-arm subset (33 of 50 tasks whose expert program moves one arm at a time), against 46.0\% for $\pi_{0.5}$ trained on RoboTwin demonstrations.
On dual-arm robots, each motion tool takes an optional \texttt{arm} argument, routed by default to the arm nearer the target.
On the 250 scenes used by PyRUA-Lean, RL improves over \model{}-SFT (21.6\% vs.\ 16.8\%) and cuts false finishes from 45 to 16.
PyRUA-Lean reaches 52.4\% there with the native embodiment, a RoboTwin-trained VLA tool and a collision-aware planner.

\paragraph{SimplerEnv.}
On the four WidowX visual-matching tasks (24 episodes each), \model{} succeeds in 83.3\% of episodes, close to ManiAgent with GPT-5 (86.8) and above VLAs fine-tuned on Bridge data such as PhysBrain~1.0 (80.2) and $\pi_{0.5}$ (57.1).

\paragraph{CaP-Bench.}
On Cube Lift and Cube Stack under CaP-Gym's step horizon, \model{} scores 75.0, above CaP-X with Gemini-3-Pro (73.5) and the 7B coder trained with CaP-RL on CaP-Bench (62.0); RL raises the score from 55.0 (\model{}-SFT).
Without the step cutoff, \model{} completes 39 of 40 trials.

\input{sections/appendix_robodojo}

\subsection{Failure taxonomy for zero-shot transfer}
\label{app:failtax}
We classified the first blocking failure of every failed episode (Table~\ref{tab:failtax}).
Three classes dominate:
\begin{itemize}
\item \textbf{Bimanual coordination.} On RoboTwin, tasks that require simultaneous two-arm motion succeed in 0 of 35 instances in every configuration.
\item \textbf{Embodiment reach.} On RoboDojo, 21 of 46 first blockers are inverse-kinematics or reach failures of the 6-DoF X5 arms.
\item \textbf{Running out of requests.} On RoboTwin, most episodes end at the request cap, often through re-verification of a solved scene; the round-400 checkpoint called \texttt{finish} in only 13--15 of 250 episodes. This matches the late-RL drift toward over-verification (Appendix~\ref{sec:collapse}).
\end{itemize}
In one test, a training-aligned variant of RoboTwin was not better than the official layout ($p{=}0.86$). The variant used a LIBERO-like camera, layouts moved into reach and LIBERO brightness. This suggests that viewpoint and lighting are not the main constraint there.

\begin{table}[t]
\centering\small
\caption{\textbf{First blocking failure} of failed transfer episodes (counts). Checkpoints: a checkpoint from an earlier RL run and \model{}-SFT, pooled; RoboTwin (dual Franka) and RoboDojo with earlier adapter versions.}
\label{tab:failtax}
\begin{tabular}{@{}lcc@{}}
\toprule
Failure Category & RoboTwin ($n=437$) & RoboDojo ($n=46$) \\
\midrule
Bimanual coordination required & \textbf{167} & 1 \\
Contact / force guard triggered & 149 & 4 \\
Reach / inverse kinematics (IK) & 22 & \textbf{21} \\
Perception / viewpoint occlusion & 34 & 8 \\
Request cap / looping & 19 & -- \\
Grasp failure (empty / slip) & 7 & 6 \\
Post-contact / placement slip & -- & 6 \\
Other (tracking, false finish, tool misuse) & 39 & -- \\
\bottomrule
\end{tabular}
\end{table}

\subsection{Zero-shot transfer episodes}
\label{app:transferep}
We show one successful \model{} episode from each of RoboTwin~2.0 and RoboDojo (Section~\ref{M-sec:transfer}). In both, the benchmark scored the episode as a success and \model{} ended it by calling \texttt{finish()}.
The frames are the head-camera images that \model{} requested during the episode. Each label gives the model request whose code cell took the image and the time since the episode started.
Code is verbatim except for cuts (\texttt{\# ...}), comments added by us (\texttt{\#}) and shortened receipts (\texttt{>>>}).
\texttt{go} is the policy's own helper, which previews a \texttt{move\_to} and then executes it (\S\ref{M-sec:sft}).

\paragraph{RoboTwin~2.0 (dual Franka, randomized scene).}
Instruction: \emph{``Use an arm to place the plastic cup for drinks on the light gray round grainy coaster''} (\texttt{place\_empty\_cup}; 7 model requests, 8 physical actions, 77\,s).

\begin{figure}[!htbp]
\centering
\includegraphics[width=\linewidth]{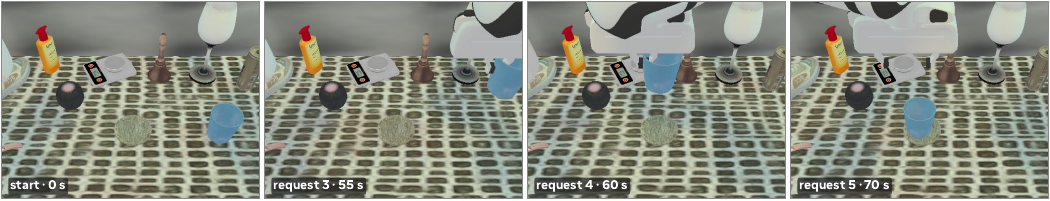}
\caption{\textbf{RoboTwin~2.0 episode.} The scene starts with distractors around the cup and coaster. \model{} grasps the cup by its rim, carries it by the measured cup-to-coaster offset, and releases it on the coaster.}
\label{fig:transferep-robotwin}
\end{figure}

{\footnotesize
\begin{verbatim}
# request 2: click the cup and the coaster in the head image, measure both
cup = robot.mark_point(obs["value"]["observation_handle"],"agentview",0.86,0.47,
                       label="blue plastic cup")
coaster = robot.mark_point(obs["value"]["observation_handle"],"agentview",0.51,0.50,
                           label="light gray round grainy coaster")
cid=cup["value"]["detection_id"]; cdid=coaster["value"]["detection_id"]
geom=robot.measure_object_geometry(cid,cdid)    # cup: cylinder, radius 4.1 cm
# request 3: grasp the +y rim at rim height, close, lift to 22 cm, observe
q=[0.0,2**-0.5,2**-0.5,0.0]
objg=geom["value"]["object"]; destg=geom["value"]["destination"]
grasp=np.array(objg["center_m"]); grasp[1]+=objg["footprint"]["radius_m"]-0.008
grasp[2]=objg["top_z_m"]-0.023
ic=robot.inspect_contact_geometry(cid,grasp.tolist(),q)
r=go(above, q, speed=0.05)
if r["success"]: r=go(grasp, q, speed=0.025, duration=25)
# ... close, lift to 22 cm, observe
# request 4: carry the cup by the measured cup-to-coaster offset
transfer=lift.copy()
transfer[:2]+=np.array(destg["center_xy_m"])-np.array(objg["center_xy_m"])
r=go(transfer, q, speed=0.06, mode="manipulation")
# requests 5-6: lower to 12 cm, open, retreat; fresh mark_point, measure_object_geometry
>>> object_to_destination_xy_offset_m=[-0.00139, 0.006599]
finish=robot.finish()    # request 7, after writing the evidence to NOTES.md
\end{verbatim}
}\par\smallskip
\model{} names both objects from the first image and measures them by clicking pixels and back-projecting them with depth. From the fitted cup radius, it places the grasp on the cup's rim rather than at its center. After the release it measures again from a fresh image. The offset of 6.7\,mm is below the 2\,cm tolerance in \texttt{TASK.md}, and \model{} calls \texttt{finish()}.

\paragraph{RoboDojo (dual ARX X5, reachable-layout variant).}
Instruction: \emph{``Stack the three blocks with different textures''} (\texttt{stack\_blocks}, layout 1; 20 model requests, 21 physical actions, 413\,s).

\begin{figure}[!htbp]
\centering
\includegraphics[width=\linewidth]{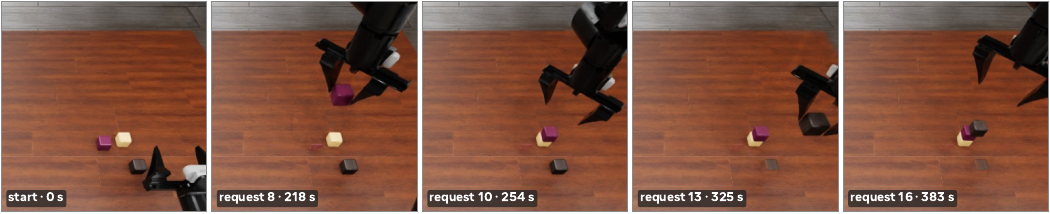}
\caption{\textbf{RoboDojo episode} (head camera, cropped to the right half of the table). \model{} lifts the purple block, puts it on the yellow block, then picks up the dark gray block and puts it on top.}
\label{fig:transferep-robodojo}
\end{figure}

{\footnotesize
\begin{verbatim}
# request 4: top-down descent to the purple block
p=np.array(objp["value"]["object"]["center_m"]); p[2]=objp["value"]["object"]["top_z_m"]-.008
r=go(p, q, speed=0.04, duration=30, angular=0.4)
>>> move_to: FAILED controller_rejected; emergency_limit; max_observed_force_n=112.1
# request 6: back off 18 mm along the executed path
args=dict(max_distance_m=.018,max_duration_s=8.)
pr=robot.preview_motion("recover",args)
rr=robot.recover(**args) if pr["success"] else pr
# request 7: re-measure, inspect pad contact, close -> jaws close empty
# request 8: reopen, rotate the gripper 45 deg about the vertical, descend slower
r=robot.move_to(gripper="open")
q2=[0,math.cos(np.pi/8),math.sin(np.pi/8),0]
grasp=np.array(pg3["value"]["object"]["center_m"]); grasp[2]=pg3["value"]["object"]["top_z_m"]-.009
r=go(above, q2, speed=0.035, duration=30, angular=0.3)
if r["success"]: r=go(grasp, q2, speed=0.015, duration=30, angular=0.3)
# ... close, lift to 23 cm, observe: block held
# requests 9-10: measure the yellow block, carry over it, lower, open, retreat
transfer=lift.copy();transfer[:2]=dest[:2];transfer[2]=.16
place=transfer.copy();place[2]=.10
# requests 11-14: verify the pair, pick the dark gray block with q2, place it on purple
# requests 15-19: fresh mark_point on all three blocks, pairwise measurements
gd9=robot.measure_object_geometry(d9["value"]["detection_id"],p9["value"]["detection_id"],
                                  shape="box",cameras="primary")
fin=robot.finish()
\end{verbatim}
}\par\smallskip
The first guarded descent stops 23\,mm short of the grasp depth on the arm's force limit. \model{} backs off with \texttt{recover}, and a closure test at the new height closes on nothing. It then reopens the gripper, rotates it by 45$^\circ$, descends more slowly, and lifts the block. It repeats this grasp for the dark gray block. Before calling \texttt{finish()}, it checks the final stack in fresh images and measures the block tops at 7.0\,cm (purple) and 10.5\,cm (dark gray), one and two block heights above the 3.5\,cm yellow block.

\section{Real robots}
\label{app:grp-real}

\input{sections/appendix_real_extra}

\input{sections/appendix_so101}

\input{sections/appendix_so101_force}

\input{sections/appendix_so101_failures}

\subsection{XR1 episode traces}
\label{app:realtrace}
Tables~\ref{tab:realtrace-a} and~\ref{tab:realtrace-b} list, request by request, two XR1 episodes, reconstructed from the \harness{} call logs of \model{}: ``Put the yellow cube in the wooden tray'' and the episode of Figure~\ref{M-fig:real}b.

\begin{table}[!htbp]
\centering
\small
\caption{\textbf{Real-robot trace (a).} Instruction in \texttt{TASK.md}: ``put the yellow cube in the wooden tray''. Times are seconds from episode start in the log and include model inference; the video in Figure~\ref{M-fig:real} is shortened, so its time stamps differ.}
\label{tab:realtrace-a}
\begin{tabular}{@{}l l p{0.41\linewidth} p{0.33\linewidth}@{}}
\toprule
Request & Time (s) & Action & Outcome \\
\midrule
1 & 7.7 & \texttt{observe()}, \texttt{robot\_geometry()} & Head and wrist images; cube lies just outside the tray's right edge \\
2 & 11.0--15.5 & \texttt{mark\_point} on cube and tray in head image; \texttt{measure\_object\_geometry} & Cube mask leaks into background (height 0.7\,mm); not used \\
3--5 & 19.0--27.2 & \texttt{segment("yellow cube")}, \texttt{segment("wooden tray")}; \texttt{select\_detection} bound to clicked points; re-measure & First binding rejected (point outside mask); after a new click, cube 3.0\,cm tall and tray interior clear \\
6 & 31.7 & \texttt{inspect\_contact\_geometry}, \texttt{preview\_motion} & Top-down grasp with cube between the jaws \\
7 & 34.9--43.2 & \texttt{move\_to} 11\,cm above the cube; \texttt{observe} & \texttt{target\_reached}; wrist view shows cube below gripper \\
8 & 45.9--52.2 & \texttt{move\_to} down at 2.5\,cm/s; \texttt{observe} & \texttt{target\_reached}, no contact stop \\
9 & 54.2 & \texttt{move\_to(gripper="close")} & Open fraction 1.00$\to$0.33; cube between fingers \\
10 & 60.2--65.0 & Lift 11.5\,cm; \texttt{observe} & Cube rises with gripper; table spot empty \\
11 & 68.5--73.4 & \texttt{move\_to} above tray center & 12\,cm transfer, \texttt{target\_reached} \\
12--16 & 76.3--98.7 & \texttt{inspect\_release\_geometry}; re-select cube (wrist) and tray (head); measure & First query rejected as stale; release projection and occluded masks judged unreliable \\
17 & 101.3--105.5 & Raise 6\,cm for clearance; \texttt{observe} & Cube still held \\
18 & 111.9--114.5 & Measure held cube in wrist view & Visible face about 3.3\,cm above table, consistent with a held cube \\
19 & 117.4--127.7 & Lower to about 1\,cm above tray floor at 2\,cm/s & 17\,cm descent, \texttt{target\_reached} \\
20 & 131.4 & \texttt{move\_to(gripper="open")} & Open fraction 0.33$\to$1.00; cube in tray \\
21 & 136.3--144.4 & Retreat 17\,cm; \texttt{hold(1\,s)}; \texttt{observe} & Gripper clear of tray \\
22 & 148.7--152.5 & Fresh \texttt{mark\_point} on cube and tray; \texttt{measure\_object\_geometry} & Center inside tray footprint, resting height consistent, offset (2.0, $-$2.1)\,cm \\
23 & 157.2 & \texttt{finish()} & Accepted; episode ends \\
\bottomrule
\end{tabular}
\end{table}

\begin{table}[!htbp]
\centering
\small
\caption{\textbf{Real-robot trace (b).} Instruction in \texttt{TASK.md}: ``Put the cube that has the color of the sky into the container that is not round.'' Times as in Table~\ref{tab:realtrace-a}.}
\label{tab:realtrace-b}
\begin{tabular}{@{}l l p{0.41\linewidth} p{0.33\linewidth}@{}}
\toprule
Request & Time (s) & Action & Outcome \\
\midrule
1 & 7.8 & \texttt{observe()}, \texttt{robot\_geometry()} & Head and wrist images \\
2 & 10.4--14.0 & Reasoning: the pale-blue cube matches the sky; the brown tray is rectangular and the white bowl is round, so it is excluded. \texttt{mark\_point} on both; \texttt{measure\_object\_geometry} & Cube 2.7\,cm tall; tray center 11\,cm away \\
3 & 18.1--18.2 & Re-measure with \texttt{shape="box"}; \texttt{inspect\_contact\_geometry} & Cube footprint 3.4$\times$3.5\,cm; centered top-down grasp \\
4 & 22.4--35.7 & \texttt{move\_to} 12\,cm above cube; guarded descent at 2.5\,cm/s; \texttt{observe} & Approach \texttt{target\_reached}; descent stops (\texttt{motion\_stalled}) about 4\,mm above target with no force; cube between open jaws \\
5--7 & 39.8--56.6 & Re-select cube (stale detection rejected; wrist camera has no depth, so head camera); \texttt{inspect\_contact\_geometry}; close; lift 7\,cm & Open fraction 1.00$\to$0.33; cube follows gripper \\
8--13 & 60.6--86.3 & Measure held cube and tray; raise 10\,cm more & Occluded masks of the held cube rejected; cube still held \\
14 & 88.6--93.2 & \texttt{move\_to} above tray center & 12\,cm transfer, \texttt{target\_reached} \\
15 & 95.4--104.7 & Lower toward tray floor at 2\,cm/s & Stops (\texttt{motion\_stalled}) about 1\,cm above endpoint, no force; cube over tray floor \\
16 & 110.5--118.8 & \texttt{move\_to(gripper="open")}; retreat 15\,cm; \texttt{observe} & Open fraction 0.33$\to$1.00; cube resting in tray \\
17--27 & 125.1--172.2 & Fresh final measurement: \texttt{mark\_point}, \texttt{segment("floor of the brown rectangular tray")}, \texttt{select\_detection} & Several tray masks rejected (table pixels, point--mask mismatch); final: center inside footprint and volume, resting height consistent, offset ($-$1.4, 3.6)\,cm \\
28 & 175.4 & Re-measure as box & Same result \\
29 & 176.9 & \texttt{finish()} & Accepted; episode ends \\
\bottomrule
\end{tabular}
\end{table}

\input{sections/appendix_speed}

%% file: sections/appendix_method_extra.tex
\subsection{Harness details}
\label{app:harness-details}

\paragraph{Context management.}
A long episode does not fit in the context window of a small model, so the harness manages the history explicitly.
Tool outputs are truncated to a fixed token budget, at most 20 receipt lines are printed per cell, and only the 12 most recent images are kept.
When the history approaches the context limit, the harness compacts it turn by turn.
It keeps the initial task context and the most recent turns verbatim, and replaces the removed turns with a short notice that lists the commands they ran and warns that the scene may have changed since.
Because the agent's variables live in the Python session rather than in the prompt, detections, poses and helper functions survive compaction.
Compaction behaves identically during RL rollouts and at test time, so the policy is trained on the inputs it will see.

\paragraph{A persistent Python session.}
The only tool is \texttt{python}, a sandboxed Python session with a pre-imported \texttt{robot} module.
Variables persist across requests, so the agent can store detections, poses and its own helper functions and reuse them later.
Robot calls do not raise exceptions; each returns a structured result with a success flag, a failure category, a message and a value.
This lets the agent write ordinary control flow (\texttt{if}, loops, retries) around robot calls within a single request.

\paragraph{Portability.}
The policy-facing API is unchanged across LIBERO, RoboTwin~2.0, RoboDojo, SimplerEnv, CaP-Bench and the real robot; backend adapters, some of which add benchmark-specific safeguards, are listed in Appendix~\ref{app:adapt}.

\subsection{One RL round}
\label{app:rlround}

Algorithm~\ref{alg:rl} summarizes one RL round of \method{} (\S\ref{M-sec:rl}).

\begin{algorithm}[h]
\caption{One RL round of \method{}}
\label{alg:rl}
\begin{algorithmic}[1]
\STATE Sample 64 start states; roll out each $G{=}4$ times with $\pi_\theta$ in \harness{}, recording the hidden snapshot $s_t$ before every request
\STATE Compute per-request credit $\delta_t$ with the current critic $V_\phi$ (Eq.~\ref{M-eq:td}) and advantages $A_{i,t}$ (Eq.~\ref{M-eq:adv})
\STATE Update the critic on all request states of the round with $\mathcal{L}_V$
\STATE Accumulate the policy gradient of $\mathcal{L}_\pi$ on up to 8 sampled requests per episode
\STATE Every 4 rounds, take an optimizer step on $\theta$ and synchronize the weights to the rollout servers
\end{algorithmic}
\end{algorithm}

%% file: sections/appendix_interp.tex
\subsection{What the policy represents}
\label{app:interp}

We ask what \model{} reads from its inputs and what its hidden states encode.
We compare \model{} (the RL checkpoint at round 300) with its SFT initialization, \model{}-SFT.
Unless stated otherwise, the data are 478 trained requests from 239 episodes of the RL run (rounds 260--300, 40 LIBERO tasks).
We score both models on the same token sequences.
We compare the two models through paired differences on identical inputs.
Brackets are 95\% confidence intervals from a bootstrap over episodes.

\paragraph{Measure.}
For a token class (e.g., the pixel coordinates of \texttt{mark\_point} calls), we report
$\Delta = \log p(\text{token}\mid\text{true input}) - \log p(\text{token}\mid\text{modified input})$
in nats per token, averaged over the tokens of that class in a request.
A large $\Delta$ means the tokens depend on the part of the input we modified.

\begin{figure}[t]
\centering
\includegraphics[width=\linewidth]{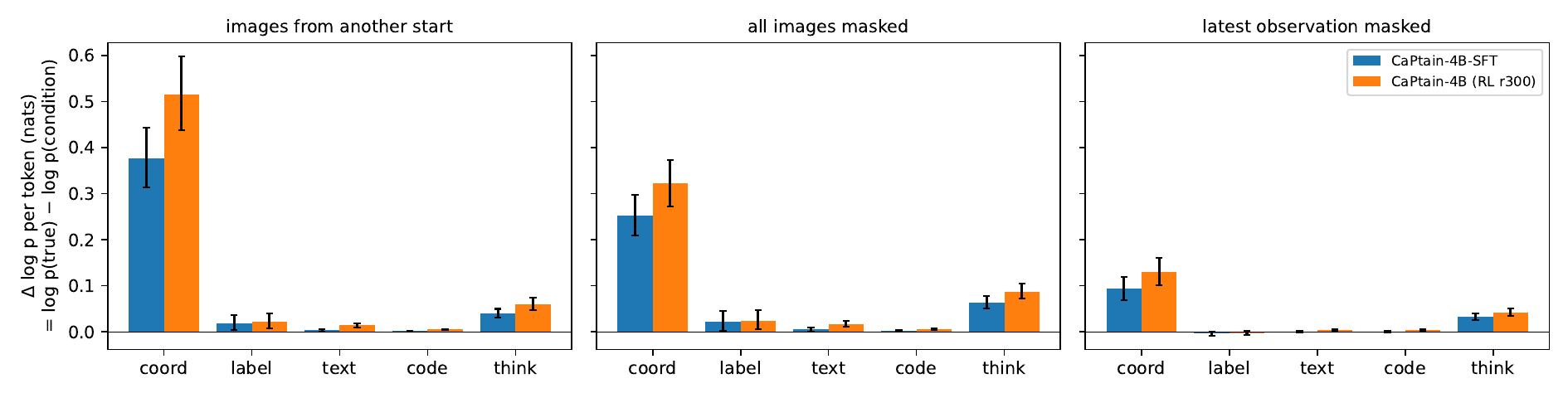}
\caption{\textbf{E1: visual grounding.} $\Delta\log p$ per token (nats) by token class when the images come from another start of the same task (left), when all images are masked (middle), and when only the latest observation is masked (right). Coordinates (\texttt{coord}) of \texttt{mark\_point} calls depend on the images; object labels and code do not. Error bars: 95\% bootstrap CI over episodes. 478 trained requests from rounds 260--300 (229 with coordinates).}
\label{fig:interp-e1}
\end{figure}

\paragraph{E1: coordinates are grounded in the image.}
We replace the images in the context with images from another start of the same task, or mask them (\cref{fig:interp-e1}).
With swapped images, the \texttt{mark\_point} coordinates become 0.515 [0.437, 0.598] nats/token less likely under \model{} and 0.377 [0.313, 0.444] under \model{}-SFT.
The paired increase from RL is +0.139 [0.116, 0.163].
Code tokens other than coordinates move by only 0.005 (\model{}) and 0.002 (\model{}-SFT).
Masking all images gives a smaller effect on coordinates (0.322 for \model{}, 0.253 for \model{}-SFT).
Masking only the latest observation gives 0.130 for \model{}, so most of the evidence also sits in earlier observations kept in context.

\begin{figure}[t]
\centering
\includegraphics[width=0.8\linewidth]{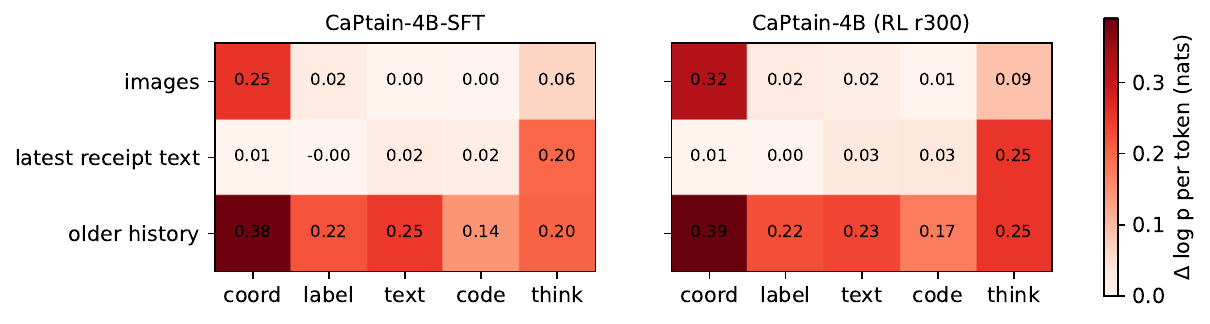}
\caption{\textbf{E3: input attribution by occlusion.} $\Delta\log p$ per token (nats) when one input segment is removed (rows) for each output token class (columns), for \model{}-SFT (left) and \model{} (right). Same 478 requests as \cref{fig:interp-e1}.}
\label{fig:interp-e3}
\end{figure}

\paragraph{E3: what each output reads.}
We remove one input segment at a time: the images, the text of the latest receipt, or the older history (\cref{fig:interp-e3}).
Coordinates depend on the images and the older turns, and not on the latest receipt text (0.01 for both models).
The reasoning text reads the latest receipt (0.20 for \model{}-SFT, 0.25 for \model{}).
RL increases this dependence: the paired change is +0.054 [0.046, 0.062] nats/token for reasoning and +0.014 [0.011, 0.016] for code.

\paragraph{E2: state probes.}
We fit linear probes on the hidden state at the generation position of each request.
The targets are the privileged quantities the \critic{} reads.
For each target we report the best layer.
Task and scene state are linearly decodable well beyond what task identity and time explain.
For \model{}, whether the target object is held is decoded at AUC 0.956, against 0.82 for a probe on task identity and request index.
The log end-effector--target distance is decoded at $R^2$ 0.413, against 0.01 for the same baseline.
RL changes these probes only slightly (distance $R^2$ +0.058 [0.026, 0.098]).
Masking every image leaves the physical-state probes unchanged (held 0.956, distance 0.431).
The physical state is therefore read mostly from the numeric receipts in context, not from the pixels.

The probes also support the same-$t$ design of the \critic{}.
The request index alone predicts eventual success at AUC 0.67, because successful episodes end with a long already-solved suffix.
The request index is itself decodable from the hidden state ($R^2$ 0.69 at layer 12).
A critic that reads the policy state would therefore see time.
Comparing predicted success at a fixed request index removes this shortcut.

\paragraph{Perception self-monitoring.}
We ask whether the policy doubts its own measurements when they are wrong.
The data are 17 episodes from RL rounds 260--300.
We take every successful \texttt{measure\_object\_geometry} result that the next request reacts to (274 results).
We label a result bad if it measures the wrong object or is off by more than 2\,cm, using privileged object positions.
An LLM annotator labels, from the next request, whether the policy used, doubted, or rejected the result.
The policy doubts or rejects 57 of 85 bad results (67\%) and 13 of 189 good results (7\%).
Doubt separates bad from good results with AUC 0.87.
On results the tool did not flag, the policy doubts 31 of 52 bad ones (60\%) and 7 of 169 good ones (4\%).
The tool flag alone catches 29 of 85 bad results.

\paragraph{Summary.}
The policy grounds its coordinates in the current image, encodes the physical state of the scene, and notices many of its own bad measurements.

%% file: sections/appendix_exp_extra.tex
\subsection{LIBERO breakdown and inference cost}
\label{app:expextra}
\begin{table*}[h]
\centering\small
\setlength{\tabcolsep}{4pt}
\caption{\textbf{SFT vs.\ RL across the LIBERO benchmark family.}
We report host and strict success rates (\%) and the paired RL gain $\Delta$ (in percentage points) over \model{}-SFT across $n$ evaluation episodes per suite.
Exact McNemar $p$-values are in Appendix~\ref{app:results} ($p < 10^{-5}$ on all primary suites).}
\label{tab:main}
\begin{tabular*}{\linewidth}{@{\extracolsep{\fill}}l r cccc ccc@{}}
\toprule
 & & \multicolumn{4}{c}{Host Success (\%)} & \multicolumn{3}{c}{Strict Success (\%)} \\
\cmidrule(lr){3-6} \cmidrule(l){7-9}
Benchmark & $n$ & Base & SFT & \textbf{\model{}} & $\Delta$ & SFT & \textbf{\model{}} & $\Delta$ \\
\midrule
LIBERO & 240 & 0.0 & 55.4 & \textbf{73.8} & +18.3 & 52.1 & \textbf{67.5} & +15.4 \\
LIBERO-PRO (3 suites) & 600 & 0.3 & 59.5 & \textbf{78.5} & +19.0 & 56.7 & \textbf{71.5} & +14.8 \\
LIBERO-PRO (All 4 suites) & 800 & 0.2 & 49.2 & \textbf{68.6} & +19.4 & 45.1 & \textbf{58.9} & +13.8 \\
LIBERO-Plus & 700 & 0.0 & 48.6 & \textbf{66.1} & +17.6 & 44.6 & \textbf{59.0} & +14.4 \\
LIBERO-PRO (4 categories) & 160 & 0.0 & 48.8 & \textbf{68.8} & +20.0 & 48.1 & \textbf{60.6} & +12.5 \\
LIBERO-MAX Dynamic & 80 & 0.0 & 40.0 & \textbf{61.2} & +21.2 & 40.0 & \textbf{50.0} & +10.0 \\
LIBERO-MAX Base & 80 & 0.0 & 51.2 & \textbf{70.0} & +18.8 & 45.0 & \textbf{63.8} & +18.8 \\
\bottomrule
\end{tabular*}
\end{table*}

\begin{figure}[h]
\centering
\includegraphics[width=\linewidth]{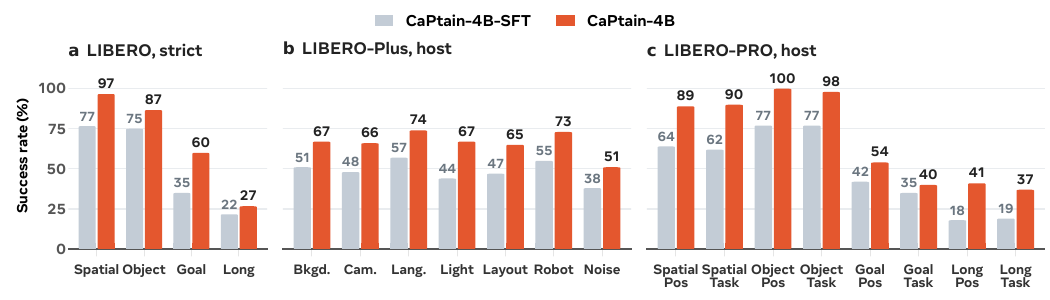}
\caption{\textbf{Breakdown}, \model{}-SFT vs.\ \model{}. (a) LIBERO strict success by suite. (b) LIBERO-Plus host success by perturbation dimension. (c) LIBERO-PRO host success by suite and perturbation type.}
\label{fig:breakdown}
\end{figure}

\paragraph{Cost.}
\label{app:cost}
A \model{} episode on LIBERO uses on average 32.6 requests (median 26), 16.6 physical actions and 7.3k output tokens, and takes 417\,s of wall time on a GPU shared by eight episodes.
This amounts to about 52 GPU-seconds of policy compute, or an estimated \$0.03--0.06 per episode at \$2--4 per GPU-hour, excluding perception services.
Published code agents report \$0.18--0.24 (SPARK, WAA, RACaP) to \$8--22 (OpenRUA) per episode.
Appendix~\ref{app:fullcompare} compares \model{} with published VLAs, world-action models and code agents on all four LIBERO-family benchmarks.

%% file: sections/appendix_robodojo.tex
\subsection{RoboDojo}
\label{app:robodojo}

\begin{table}[h]
\centering\footnotesize
\caption{\textbf{RoboDojo-Sim} (dual 6-DoF ARX X5), zero-shot. Official leaderboard Score and success rate (SR, \%).}
\label{tab:robodojo}
\begin{tabular}{@{}l l rr@{}}
\toprule
Method & Backbone / Type & Score & SR \\
\midrule
RoboDojo Agent~\citep{astra2026robodojo} & GPT-6 Astra & 29.0 & 22.5 \\
RoboDojo Agent~\citep{astra2026robodojo} & DeepSeek-Flash & 3.0 & 1.9 \\
RoboDojo Agent~\citep{astra2026robodojo} & GPT-5.5 & 1.1 & 0.9 \\
RDT~\citep{liu2024rdt} & In-domain VLA & 0.5 & 0.1 \\
OpenVLA-OFT~\citep{kim2025oft} & In-domain VLA & 0.2 & 0.0 \\
\rowcolor{oursbg} \textbf{\model{} (ours, all 42)} & \textbf{Qwen3.5-4B} & \textbf{1.7} & \textbf{0.8} \\
\rowcolor{oursbg!45} \model{} (ours, 1-arm 32) & Qwen3.5-4B & 2.0 & 1.0 \\
\bottomrule
\end{tabular}
\end{table}

RoboDojo tests zero-shot cross-embodiment transfer from LIBERO's single 7-DoF Panda to dual 6-DoF ARX X5 arms across 42 tasks.
When the \texttt{arm} argument is omitted, the harness routes a call to the arm whose base is nearer the target, unless the active arm holds an object; a \texttt{move\_both} tool moves both arms together.
The single-arm subset contains the 32 tasks whose specification requires no simultaneous two-arm action, handover or navigation.
We compare against the RoboDojo agent driven by GPT-6 Astra, DeepSeek-Flash or GPT-5.5, and two VLAs (RDT and OpenVLA-OFT) fine-tuned on the RoboDojo training set.
GPT-6 Astra reaches a Score of 29.0; the other frontier models (DeepSeek-Flash 3.0, GPT-5.5 1.1) and the in-domain VLAs (RDT 0.5, OpenVLA-OFT 0.2) score low.
Without RoboDojo data, \model{} reaches a Score of 1.7 (0.8\% success; 2.0 and 1.0\% on the single-arm tasks) and solves tasks in the Generalization and Open splits.
Its main bottleneck is kinematic reach: the top-down workspace of the 6-DoF arm extends only ${\approx}0.5$\,m from its base (Appendix~\ref{app:failtax}).

%% file: sections/appendix_real_extra.tex
\subsection{Simulated twins for real-robot adaptation}
\label{app:realextra}

\begin{figure}[h]
\centering
\includegraphics[width=\linewidth]{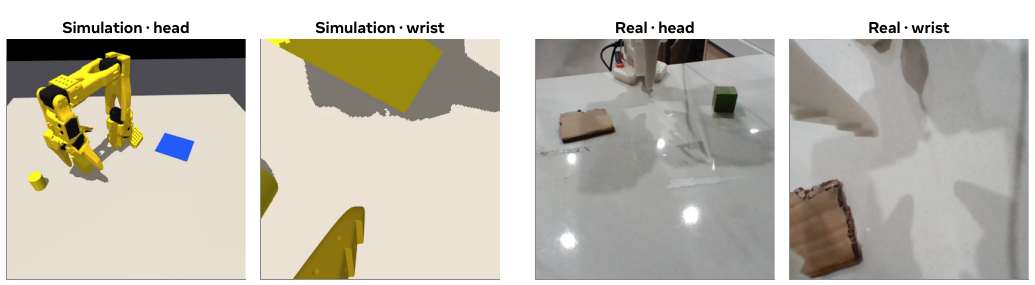}
\caption{\textbf{SO-101: simulated twin used for adaptation} (left: head and wrist camera at the start of a training episode) \textbf{and the real scene at test time} (right: the first head and wrist images of the pick-and-place episode ``Please place the green block on the brown mat.'').}
\label{fig:twin}
\end{figure}

Each real arm is paired with a simulated twin of its kinematics and gripper, and \model{} is adapted on episodes collected in that twin only.
The twin uses primitive objects and plain surfaces and does not reconstruct the real scene (Figure~\ref{fig:twin}); it teaches the policy the conventions of the new embodiment, such as its reach, gripper geometry, and camera placement.
On the real arm, the backend replaces the simulator with the arm, its head and wrist cameras, a kinematic model updated from the measured joints, and safety guards; the policy-facing API and receipts are unchanged.
The XR1 of Figure~\ref{M-fig:real}b is adapted in the same way with a twin of the XR1.

%% file: sections/appendix_so101.tex
\subsection{SO-101 real-robot episodes}
\label{app:so101}
\begin{figure}[p]
\centering
\newcommand{\soepisode}[2]{\textbf{#1}\par\smallskip\includegraphics[width=\linewidth]{figs/so101/#2.jpg}\par\medskip}
\soepisode{(a) Pick: ``Pick the green cube for 10 seconds''}{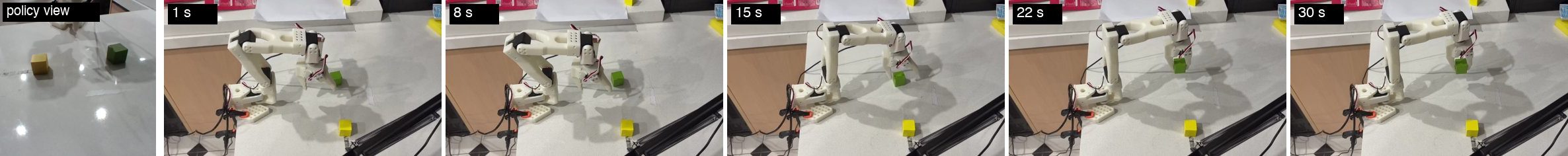}
\soepisode{(b) Pick \& place: ``Place the yellow square into the blue square on the white paper.''}{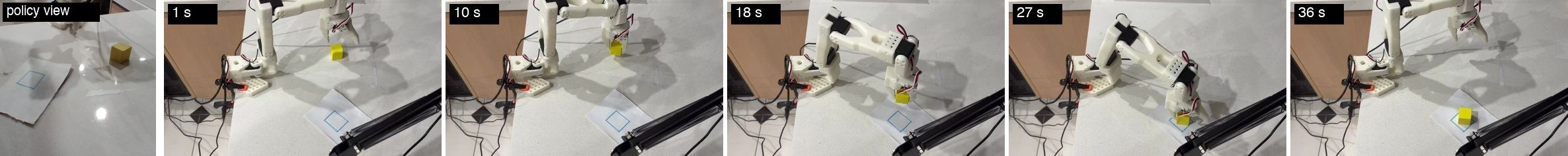}
\soepisode{(c) Pick \& place: ``Place the yellow square into the blue square on the white paper.''}{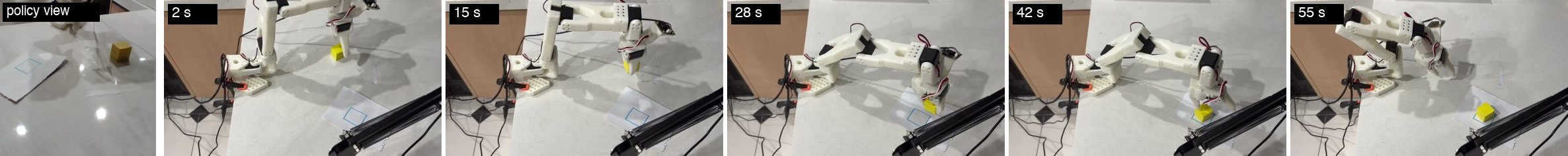}
\soepisode{(d) Pick \& place: ``Please place the green block on the brown mat.''}{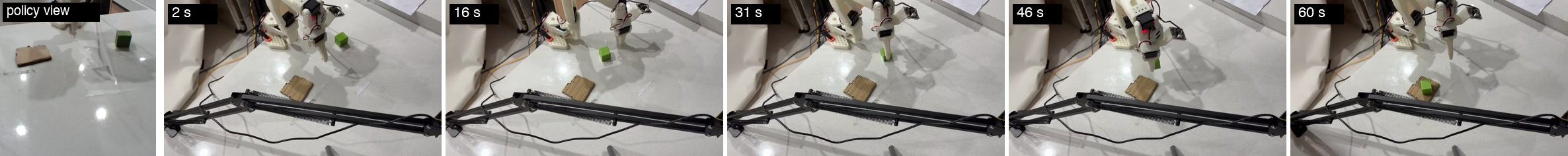}
\soepisode{(e) Stack: ``Pick the green cube and stack it on yellow cube''}{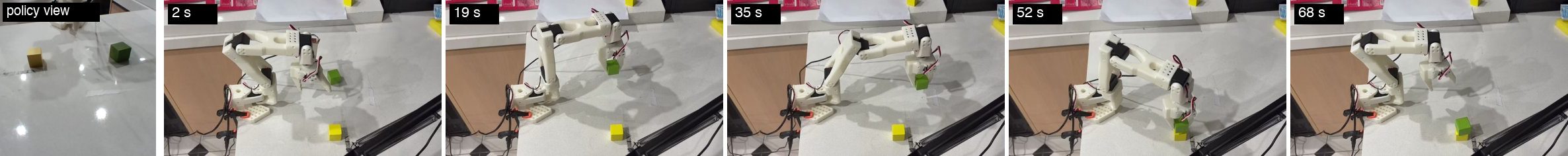}
\soepisode{(f) Stack: ``Pick the green cylinder and stack it on yellow cube''}{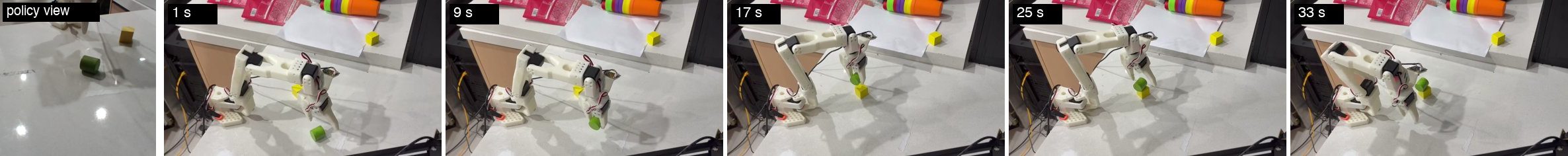}
\soepisode{(g) Ordered placement: ``First, place the yellow block on the plate, then place the green block on the plate, and do not stack the two blocks.''}{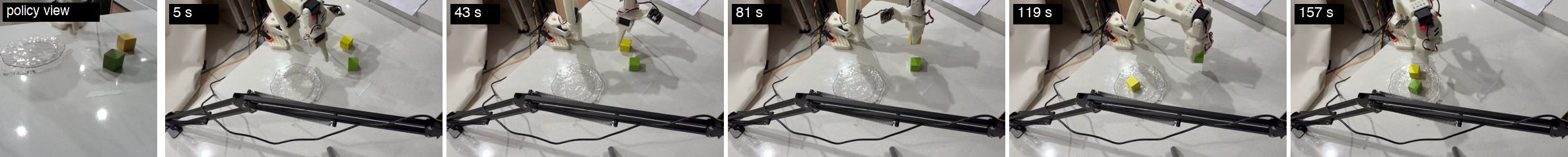}
\soepisode{(h) Commonsense: ``Please select the only meat item shown in the picture and place it in the bowl.''}{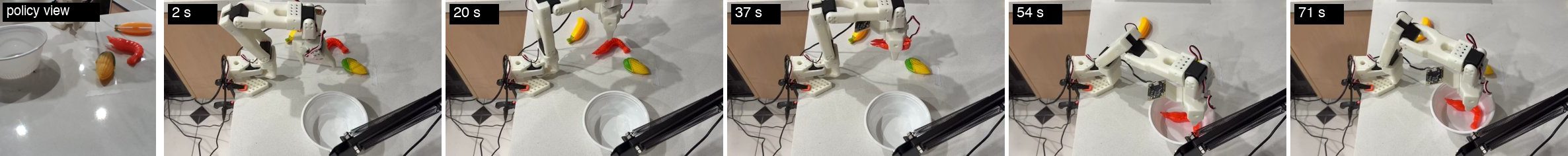}
\caption{\textbf{SO-101 episodes.} Left: the first head-camera image the policy receives. Right: five frames of the episode video, with the video time. In (f) the object to stack is a cylinder lying on its side; in (g) the agent places the yellow block first and keeps the two blocks apart; in (h) it identifies the shrimp as the only meat item among three toy ingredients and places it in the bowl.}
\label{fig:so101eps}
\end{figure}

This subsection shows SO-101 episodes of \model{} for the tasks of Table~\ref{M-tab:real} and for an instruction that requires commonsense reasoning about objects.
The policy is the same in every episode, the instruction is the only input that changes, and the setup follows Section~\ref{M-sec:real}: a fixed head camera and a wrist camera, the unchanged policy-facing API of \harness{}, and no real-world training data.
Table~\ref{tab:so101eps} lists the instructions verbatim together with the matching video on the project page (\url{https://waltstephen.github.io/captain/}), and Figure~\ref{fig:so101eps} shows the first head-camera image the policy receives (left) and five frames of each video.
Videos are recorded by a third-person camera and have no audio.

\begin{table}[h]
\centering\small
\caption{\textbf{SO-101 episodes shown on the project page.} Requests are model calls; actions are executed physical actions; time is the wall time of the episode.}
\label{tab:so101eps}
\begin{tabularx}{\linewidth}{@{}l>{\raggedright\arraybackslash}Xlccc@{}}
\toprule
Task & Instruction & Video & Requests & Actions & Time (s) \\
\midrule
Pick & ``Pick the green cube for 10 seconds'' & \texttt{01\_pick.mp4} & 10 & 6 & 47 \\
Pick \& place & ``Place the yellow square into the blue square on the white paper.'' & \texttt{02\_pick\_place\_paper\_a.mp4} & 19 & 12 & 65 \\
Pick \& place & ``Place the yellow square into the blue square on the white paper.'' & \texttt{03\_pick\_place\_paper\_b.mp4} & 14 & 9 & 57 \\
Pick \& place & ``Please place the green block on the brown mat.'' & \texttt{04\_pick\_place\_mat.mp4} & 20 & 14 & 73 \\
Stack & ``Pick the green cube and stack it on yellow cube'' & \texttt{05\_stack\_cube.mp4} & 18 & 16 & 69 \\
Stack & ``Pick the green cylinder and stack it on yellow cube'' & \texttt{06\_stack\_cylinder.mp4} & 8 & 8 & 43 \\
Ordered placement & ``First, place the yellow block on the plate, then place the green block on the plate, and do not stack the two blocks.'' & \texttt{07\_ordered\_placement.mp4} & -- & -- & -- \\
Commonsense & ``Please select the only meat item shown in the picture and place it in the bowl.'' & \texttt{08\_meat\_selection.mp4} & 37 & 13 & 90 \\
\bottomrule
\end{tabularx}
\end{table}

%% file: sections/appendix_so101_force.tex
\subsection{Contact and force estimation without a force sensor}\label{app:so101force}

The simulated arms that \model{} is trained on report a wrist force, and the contact-aware primitives of \harness{} rely on it: a guarded descent stops at first contact, a closing gripper reports whether it met an object, a release is checked for unloading, and every action receipt states the peak force seen and why the motion stopped.
The SO-101 used in Section~\ref{M-sec:real} has no force/torque sensor.
We therefore estimate the external force on the hand from the servos themselves and feed this estimate into exactly the same detectors, with the same thresholds and the same receipts as in simulation.
This subsection describes the estimator, its per-episode calibration, and how each primitive uses it.

\paragraph{From servo tracking error to a hand force.}
Each joint of the SO-101 is a low-cost position-controlled servo, which behaves approximately as a proportional spring around its goal position.
When the hand presses against something, the joints that carry the load fall behind their goals, so the position error is a proxy for joint torque.
The arm is commanded at 20\,Hz, and each control-step goal is spread over five intermediate servo goals so the servos follow a ramp.
At every control step the estimator reads the measured joint positions and compares them not with the latest goal but with the span of goals sent during the last six steps (300\,ms): a servo that simply lags a moving goal stays within that span and contributes nothing, whereas a joint that is blocked falls out of the span as the goals move on.
Gravity sag is handled before this comparison: the measured joints are corrected with a model of the arm's gravity load, and an integral trim, active while a joint's goal is still, removes the remaining steady-state error.
Errors below a deadband of 50 encoder steps are treated as friction and residual gravity error and ignored.
The excess error is converted to a joint torque with a nominal servo stiffness of 20\,N\,m/rad and mapped to a Cartesian force at the grip point through the transpose of the calibrated arm's position Jacobian (pseudo-inverse over the five arm joints).
Near the table, a hand goal that would bring the lowest fingertip below 3\,mm under the table plane is not sent; the refused depth is added to the vertical force through a stiff virtual spring of 2000\,N/m, so pushing into the table reads as an upward contact force even though the arm does not move.
Finally, as in the simulated driver, forces below 0.6\,N are reported as zero.
The result is a three-dimensional force in the robot base frame that the driver reports in place of a wrist sensor reading.

\paragraph{Held objects and placing.}
The jaw is handled separately.
When a closing jaw stops moving (less than 4 encoder steps over three consecutive control steps) before it reaches its commanded position, it has met an object; the jaw goal is then set only 120 steps past the stalled position, so the object is held with a light squeeze instead of the servo's full torque.
A jaw that stalls more than 60 steps wider than its fully closed position (about 2100 versus 2040 raw steps) therefore indicates an object between the jaws.
While such an object is held and the hand is being lowered, a support under the object is felt only through the small extra load it adds, so the deadband on the arm joints is reduced from 50 to 20 encoder steps.
This is what lets a placement stop on top of a lower block instead of pushing it aside.

\paragraph{Per-episode calibration.}
Before each episode the arm traces a small horizontal triangle in free space (legs of 15\,mm), after a short stationary hold, and the force estimate is recorded throughout (20 stationary and 140 moving samples in the logged episodes).
From these samples \harness{} computes, exactly as it does for a simulated wrist sensor, a force bias (median during the hold), the noise scale of the force magnitude during motion (median and median absolute deviation), and per-axis noise.
Contact thresholds are set to the median plus six times the noise scale, with a floor of 1\,N.
Because the deadbands absorb free-motion tracking error, the estimate stays at zero during the calibration motion on the real arm, the bias is zero, and the thresholds sit at the 1\,N floor.
The calibration also checks that the triangle returns to its start (within 2\,mm in the logged episodes).

\paragraph{How the primitives use the estimate.}
All detectors are those of the simulated harness, applied to the estimated force:
\begin{itemize}
  \item \emph{Guarded motion.} By default every arm motion is guarded: it stops as soon as the bias-corrected force exceeds the calibrated threshold (1\,N) for three consecutive samples, or immediately when it exceeds five times the threshold.
  The receipt reports the stop reason \emph{contact detected}, the distance actually travelled, and the peak force.
  Over 85 such contact stops logged on the real arm, the median reported peak was 5.9\,N.
  \item \emph{Manipulation motion.} For motions where contact is expected (pushing, sliding), the policy supplies a load budget; the motion stops when the force stays above that budget for 0.25\,s, or when the hand is loaded but advances less than 2\,mm in 0.5\,s (reported as a contact stall).
  \item \emph{Stationary contact.} Outside these detectors, a contact force above the threshold that persists for 0.25\,s also stops the motion.
  \item \emph{Grasping.} While the jaw closes, the hand force is monitored against its value at the start of the closure; a rise of 5\,N (on two samples) stops the closure, which catches a jaw that is pushing the object or the table rather than enclosing it. After closing, the gripper reports an object as detected when it was commanded fully closed but its measured opening stays above 2\% of its range, i.e., the jaw stalled on something.
  \item \emph{Release.} While the jaw opens, the release is aborted if the force rises more than 2\,N above the lowest value seen during the release, if the jaw opening goes backwards, or if a force of 18\,N or more does not drop by at least 1\,N within 0.5\,s.
  \item \emph{Safety.} Any estimated force above 60\,N stops every motion, independently of mode.
\end{itemize}

\begin{table}[h]
  \centering
  \small
  \begin{tabularx}{\linewidth}{@{}lXX@{}}
    \toprule
    Signal & What it detects & What the policy sees \\
    \midrule
    Arm servo position error beyond the recent goals (50-step deadband) & Hand pressing against the table or an object & Motion stops; receipt reports \emph{contact detected}, peak force, distance travelled \\
    Same, with 20-step deadband while lowering a held object & Support under a held object during placing & Placement stops on the support; same receipt \\
    Refused goal below the table plane (virtual spring) & Hand pushed into the table & Upward contact force; guarded stop \\
    Jaw stalling before fully closed & Object between the jaws & Gripper state reports an object detected and the measured opening \\
    Force rise during closing or opening & Jaw pushing instead of grasping; object still loaded during release & Closure or release stops with the corresponding reason \\
    \bottomrule
  \end{tabularx}
  \caption{Servo-derived signals on the SO-101 and how they reach the policy.}
  \label{tab:so101force}
\end{table}

\paragraph{Same interface as in simulation.}
The policy never sees servo errors or raw encoder steps.
It receives a wrist force in the base frame, gripper state, and action receipts with the same fields as in simulation: the stop reason (e.g., target reached, contact detected, contact stall, release force rise), whether the endpoint was reached, the executed distance and displacement, the maximum observed force, and the gripper opening before and after.
The contact thresholds come from the same calibration procedure and the same constants.
Consequently \model{} uses the estimated force on the real arm exactly as it used the simulated one, through the same tools and receipts.

%% file: sections/appendix_so101_failures.tex
\subsection{SO-101 failure cases}
\label{app:so101fail}
\begin{figure}[t]
\centering
\newcommand{\failrow}[2]{\textbf{#1}\par\smallskip\includegraphics[width=0.82\linewidth]{figs/so101_fail/#2.jpg}\par\medskip}
\failrow{(a) Unreachable pose, recovered: ``Please select the only meat item shown in the picture and place it in the bowl.''}{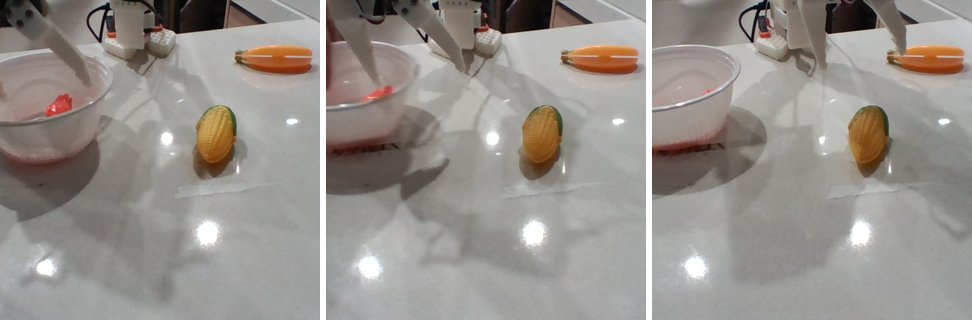}
\failrow{(b) Calibration offset, stacked cube slides off: ``Please clip the green square onto the yellow square, aligning the centers for stability.''}{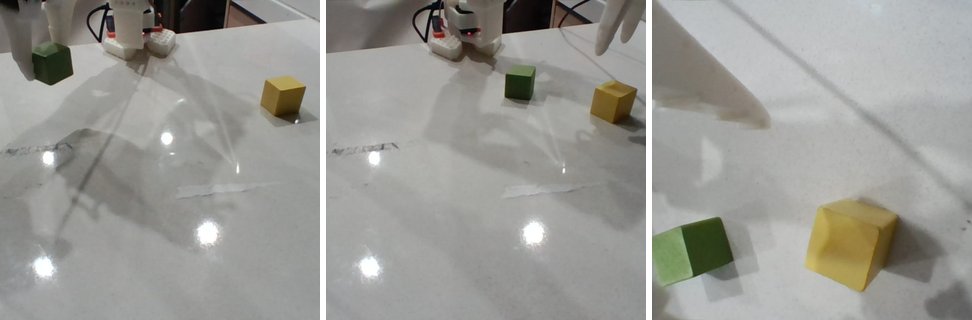}
\failrow{(c) Calibration offset, cylinder falls off at release: ``Pick the green cylinder and stack it on yellow cube''}{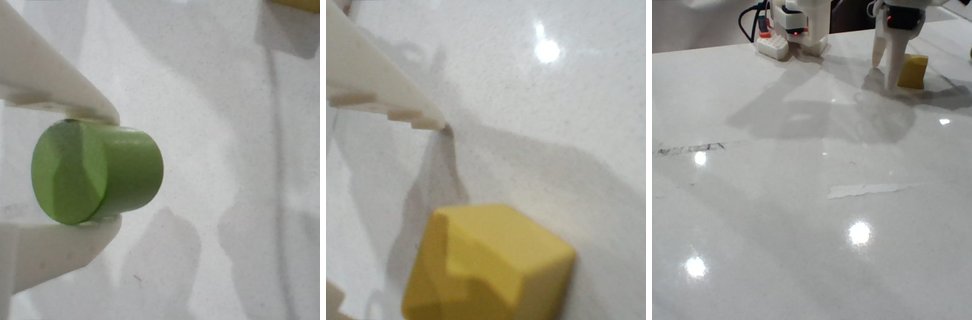}
\caption{\textbf{SO-101 failure cases}, from the images the policy receives. (a) Head camera: after the release in the bowl; at the intermediate waypoint chosen after three motions are rejected as unreachable; final check before finishing. (b) Head camera with the green cube in the gripper after the grasp; head and wrist cameras after release: the green cube rests on the table next to the tilted yellow cube. (c) Wrist camera holding the cylinder before it is carried to the cube; wrist and head cameras after opening the jaws above the cube: the cube lies at the edge of the wrist view and the cylinder is no longer on it.}
\label{fig:so101fail}
\end{figure}

This subsection analyzes where the remaining failures of \model{} on the SO-101 (Section~\ref{M-sec:real}, Table~\ref{M-tab:real}) come from.
We analyzed the call logs of all 52 real-arm episodes, together with the camera images the policy received.
Two properties of the platform account for most of the physical failures: the limited kinematics of a low-cost five-joint arm, and the millimeter-level precision of the camera-to-arm calibration.
Figure~\ref{fig:so101fail} shows one representative case of each kind and a second calibration case.

\paragraph{Kinematic limits of a five-joint arm.}
The SO-101 has five arm joints (base pan, shoulder, elbow, wrist flex, wrist roll) and a parallel gripper.
A gripper pose has six degrees of freedom, so only a five-dimensional subset of poses is reachable: the gripper axis always lies in the vertical plane through the base axis, and a tilted approach is feasible only when its tilt points along that plane.
\harness{} checks inverse kinematics at every sample of a requested straight-line motion, with the orientation interpolated between start and goal, and rejects the motion before execution if any sample is unreachable; the arm does not move and the policy receives a receipt that names the failure.
Across the 52 episodes, 229 of 1381 motion commands (17\%) were rejected this way, in 22 episodes.
Most rejected requests asked for a tilted gripper (182 of 229); the rest were top-down poses at the edge of the workspace.
Rejections cost time but are safe, and the policy usually changes its request: in the next motion it issues after a rejection, it changes the gripper orientation in 161 of 200 cases and the target position by more than 1\,cm in 123 of 189 cases.
Figure~\ref{fig:so101fail}a shows a typical recovery in the commonsense episode of Appendix~\ref{app:so101}.
After releasing the shrimp into the bowl, the policy asks to lift the gripper while turning it to a top-down orientation; the request is rejected.
Two further requests that keep the tilted grasp orientation but move straight across the table to the other side of the base are rejected as well, because the wrist cannot keep that orientation along the whole path.
The policy then sends the arm to an intermediate waypoint on the same side of the base, which is accepted, continues from there, takes a fresh image, confirms that the shrimp is in the bowl, and finishes.

\paragraph{Precision of the camera-to-arm calibration.}
The policy localizes objects in the fixed head camera, whose pose relative to the arm is calibrated once.
To quantify the residual error, we compared head-camera estimates of free cubes with the cube centers seen by the wrist camera looking straight down on them (141 views).
The head-camera estimates carry a nearly constant horizontal offset of $(-3.4, +6.7)$\,mm (39 independent placements), which is subtracted; the remaining median error is 7.1\,mm in a leave-one-episode-out evaluation.
A second source of offset is the gripper: a held cube is pushed against the fixed jaw, and its center sits 5.8\,mm (median of 20 wrist views) off the jaw center along the closing direction, which the commanded grasp point compensates with a 5.5\,mm shift.
For pick and pick-and-place these errors are small compared with the object and target sizes: in the successful episodes of Appendix~\ref{app:so101}, the policy's own final measurement places the object 6 to 13\,mm from the target center.
For stacking, the same error is a sizable fraction of a 3\,cm cube, and the upper object can land on an edge of the lower one.
In Figure~\ref{fig:so101fail}b, the policy carries the green cube above its estimate of the yellow cube and opens the jaws; the green cube lands partly on the edge, tips the yellow cube, and slides onto the table beside it.
In Figure~\ref{fig:so101fail}c, the object is an upright cylinder; the wrist image at release shows the yellow cube at the edge of the view instead of under the gripper, and the cylinder falls off when the jaws open.
In both cases the policy checks the outcome in the next image: it re-observes, re-marks the objects in the fresh image, and re-measures before acting again.
The policy re-measures from a fresh image before every grasp and placement; the remaining failures occur where the residual offset meets a small support surface.

%% file: sections/appendix_speed.tex
\section{Fast local inference}
\label{app:speed}

\paragraph{Why a local policy.}
\method{} controls a robot through a closed loop of many short model calls: each call reads the latest tool receipts and images, reasons briefly, and emits a few lines of code that perceive, move, or grasp. A successful SO-101 episode in our setting uses 7--44 such calls, so the latency of a single call is multiplied by the length of the loop and directly determines how long the arm waits idle between motions. The same loop structure makes per-call cost, network availability, and reaction time matter: a policy that runs on the robot's own GPUs incurs no API cost, does not depend on a remote service, and can respond to a change in the scene within one or two calls rather than after a long deliberation. We therefore optimize both the serving of the 4B model and the latency of the harness around it. All measurements in this section are on the simulated twin of the SO-101 arm (\cref{app:so101}) with realistic perception (RGB only, monocular depth, segmentation and pointing models; no privileged state), one episode at a time, on two RTX~4090 GPUs. Time is reported as \emph{robot wall time}: simulator wall time minus simulation-only work, plus the commanded motion time on the arm's 20\,Hz servo schedule, i.e.\ the time an episode would take on the physical arm.

\paragraph{Serving the 4B policy.}
In bf16, model computation accounts for 88\% of episode wall time, and decoding dominates it (91 output tokens/s for a single stream). We apply four lossless or near-lossless changes. (i) \emph{Weight quantization}: all linear layers are quantized to 4-bit weights with 16-bit activations and the output projection to 8-bit weights, which doubles single-stream decode throughput (91$\rightarrow$178 tokens/s) and reduces the policy weights to under 5\,GB, so that the policy and the segmentation model share one 24\,GB GPU while the pointing and depth models occupy the second. (ii) \emph{Speculative decoding with a multi-token-prediction (MTP) draft head}: we attach an MTP head to the policy and fine-tune it on \model{}'s own recorded outputs, which raises the number of accepted tokens per verification step by 30\% over the generic head; with four draft tokens the deployed system accepts 3.68 tokens per step, for 2.38\,ms per generated token. The draft head proposes from a 16k-token subset of the vocabulary, and verification steps use a split-KV attention kernel, which reduces full-attention cost of a verification step from 1.9\,ms to 0.22\,ms per layer at batch size one. (iii) \emph{Prefill}: prefill uses FlashAttention-2 kernels with an 8192-token batched-prefill budget, and the prompt is kept append-only (old images are dropped in chunks rather than one at a time) so that consecutive requests reuse the prefix cache. (iv) \emph{Bounded deliberation}: the thinking budget per call is capped at 256 tokens; together with the fine-tuned MTP-4 head this lowers model time per request from 1.04 to 0.76\,s (last row of \cref{tab:speed_serving}). The attention, vocabulary-subset, and prefix-caching changes produce greedy outputs identical to the unmodified server. \Cref{tab:speed_serving} summarizes the effect on per-request latency.

\begin{table}[h]
\centering
\small
\begin{tabular}{lcc}
\toprule
Serving configuration (realistic perception) & Model s / request & Robot-time s / request \\
\midrule
bf16 & 5.14 & -- \\
+ 4-bit weights & 2.85 & 4.14 \\
+ 8-bit output head, MTP-3 drafting, faster motion & 1.47 & 2.76 \\
+ append-only image history, perception caching & 1.31 & 2.16 \\
+ FA2 prefill, draft vocabulary subset & 1.04 & 1.70 \\
+ fine-tuned MTP-4 head, thinking budget 256 (\textbf{final}) & \textbf{0.76} & \textbf{1.22} \\
\bottomrule
\end{tabular}
\caption{Per-request latency of \model{} on the SO-101 twin with realistic perception (one episode at a time, two RTX~4090). \emph{Robot-time} includes model time, motion, perception, and observation. Each row adds to the previous one; the final system needs 6.8$\times$ less model time per request than bf16 serving.}
\label{tab:speed_serving}
\end{table}

\paragraph{Harness-side latency.}
Outside the model, \harness{} adds 5\,ms between a response and the next request, since tool code runs in a persistent sandbox. The remaining non-model cost is perception and motion. Perception calls are dominated by the segmentation model's image encoder (0.36--0.42\,s per call); \harness{} caches the encoder output per image, so repeated queries on the same observation (\eg multi-view geometry estimation after pointing) skip re-encoding, which cuts perception time per episode from 8.3\,s to 6.7\,s with identical responses. Motion commands are executed with a distance-aware velocity profile that accelerates to the arm's speed limit and decelerates near the target, and gripper commands end as soon as the measured jaw position stops changing instead of after a fixed delay. In the final system a request takes 1.22\,s of robot time on average: 0.76\,s model time, 0.27\,s motion, 0.16\,s perception and observation, and 0.03\,s tool execution and harness overhead. A successful episode takes a median of 24.0\,s of robot time (12.2--67.0\,s); by task, 16.8\,s for pick-and-lift, 25.3\,s for pick-and-place onto a mat, and 36.6\,s for placing into a container.

\paragraph{Comparison with frontier models.}
We compare against frontier models running in the Codex harness, which, like \harness{}, lets the model solve the task by writing and executing code against the robot, on the same SO-101 twin with the same tasks, starting layouts, control rate, and speed limits. We evaluate GPT-6.1-Sol at low and high reasoning effort and DeepSeek-V4.1-Flash with thinking enabled. \Cref{tab:speed_codex} reports robot time on three shared episodes that all systems complete and the median over each system's successful episodes. The frontier models complete the tasks but take 4--15$\times$ longer than \model{} in median time.

\begin{table}[h]
\centering
\small
\setlength{\tabcolsep}{4pt}
\begin{tabular}{lccccc}
\toprule
& \multicolumn{3}{c}{Robot time to success (s)} & Median over & \\
\cmidrule(lr){2-4}
System & mat & container & lift & successes (s) & s / request \\
\midrule
Codex + GPT-6.1-Sol (low) & 97.3 & 129.1 & 205.7 & 150.8 & -- \\
Codex + GPT-6.1-Sol (high) & 402.2 & 289.2 & 819.6 & 356.6 & -- \\
Codex + DeepSeek-V4.1-Flash & 70.7 & 180.4 & 68.7 & 101.0 & 2.8 \\
\midrule
\model{} (local, 2$\times$RTX~4090) & \textbf{16.5} & \textbf{18.2} & \textbf{16.0} & \textbf{24.0} & \textbf{1.22} \\
\bottomrule
\end{tabular}
\caption{\textbf{Speed comparison with frontier models in the Codex harness} on the SO-101 twin, same tasks and starting layouts. Per-task columns: one shared starting layout per task (pick-and-place onto a mat, place into a container, pick-and-lift). Latency of \model{} is robot time per request (model, motion, and perception); latency of DeepSeek-V4.1-Flash is the median model response time.}
\label{tab:speed_codex}
\end{table}

\paragraph{Reaction to disturbances.}
Speed matters most when the scene changes. In a disturbance test, the target object is moved 6--10\,cm to a new position at the moment the gripper first descends onto it (within 4\,cm horizontally and below 3.5\,cm above the object's top); the new pose is identical for all systems on a given layout. We measure the robot time from the move until the object is next between the gripper jaws. On four layouts, \model{} re-grasps the object 11.7, 13.8, 15.0, and 69.0\,s after the move, with successful episodes lasting 40.3--187.3\,s in total. In the Codex harness, DeepSeek-V4.1-Flash re-grasps after 71.9--84.5\,s and GPT-6.1-Sol at high effort after 235.5--889.5\,s, with successful episodes lasting 1043.8 and 1228.7\,s. Because \model{} issues a new short call roughly every second, it observes the change and re-plans within a few calls.